\documentclass[sn-apa,iicol,pdflatex]{sn-jnl}

\usepackage{graphicx}%
\usepackage{multirow}%
\usepackage{amsmath,amssymb,amsfonts}%
\usepackage{amsthm}%
\usepackage{mathrsfs}%
\usepackage[title]{appendix}%
\usepackage[table]{xcolor}%
\usepackage{colortbl}
\usepackage{textcomp}%
\usepackage{manyfoot}%
\usepackage{booktabs}%
\usepackage{algpseudocode}%
\usepackage{listings}%

\usepackage[utf8]{inputenc} 
\usepackage[T1]{fontenc}    
\usepackage{hyperref}       
\usepackage{booktabs}       
\usepackage{amsfonts}       
\usepackage{nicefrac}       
\usepackage{microtype}      

\usepackage{graphicx}
\usepackage{subfig}

\usepackage{bm}
\usepackage{bbm}

\usepackage{float}

\usepackage[ruled]{algorithm2e}

\usepackage[capitalize]{cleveref}
\crefname{section}{Section}{Secs.}
\Crefname{section}{Section}{Sections}
\Crefname{table}{Table}{Tables}
\crefname{table}{Tab.}{Tabs.}
\Crefname{algorithm}{Algorithm}{Algorithms}
\crefname{algorithm}{Algo.}{Algos.}

\usepackage[normalem]{ulem}

\usepackage{soul}
\useunder{\uline}{\ul}{}

\begin{document}

\title[Article Title]{AROID: Improving Adversarial Robustness Through Online Instance-Wise Data Augmentation}


\author[1]{\fnm{Lin} \sur{Li}}\email{lin.3.li@kcl.ac.uk}

\author[2]{\fnm{Jianing} \sur{Qiu}}\email{jianing.qiu17@imperial.ac.uk}

\author[1,3]{\fnm{Michael} \sur{Spratling}}\email{michael.spratling@kcl.ac.uk}

\affil[1]{\orgdiv{Department of Informatics}, \orgname{King's College London}, \orgaddress{\street{Aldwych}, \postcode{WC2B 4BG}, \state{London}, \country{UK}}}

\affil[2]{\orgdiv{Department of Computing}, \orgname{Imperial College London}, \orgaddress{\postcode{SW7 2AZ}, \state{London}, \country{UK}}}
\affil[3]{\orgdiv{Department of Behavioural and Cognitive Sciences}, \orgname{University of Luxembourg}, \orgaddress{\postcode{L-4366}, \state{Esch-Belval}, \country{Luxembourg}}}


\abstract{Deep neural networks are vulnerable to adversarial examples. Adversarial training (AT) is an effective defense against adversarial examples. However, AT is prone to overfitting which degrades robustness substantially. Recently, data augmentation (DA) was shown to be effective in mitigating robust overfitting if appropriately designed and optimized for AT. This work proposes a new method to automatically learn online, instance-wise, DA policies to improve robust generalization for AT. This is the first automated DA method specific for robustness. A novel policy learning objective, consisting of Vulnerability, Affinity and Diversity, is proposed and shown to be sufficiently effective and efficient to be practical for automatic DA generation during AT. Importantly, our method dramatically reduces the cost of policy search from the 5000 hours of AutoAugment and the 412 hours of IDBH to 9 hours, making automated DA more practical to use for adversarial robustness.
This allows our method to efficiently explore a large search space for a more effective DA policy and evolve the policy as training progresses. Empirically, our method is shown to outperform all competitive DA methods across various model architectures and datasets. Our DA policy reinforced vanilla AT to surpass several state-of-the-art AT methods regarding both accuracy and robustness. It can also be combined with those advanced AT methods to further boost robustness.
Code and pre-trained models are available at: \url{https://github.com/TreeLLi/AROID}.}




\maketitle

\section{Introduction}
Deep neural networks (DNNs) are well known to be vulnerable to infinitesimal yet highly malicious artificial perturbations in their input, i.e., adversarial examples \citep{szegedy_intriguing_2014}. The lack of robustness cause a crisis of security and trustworthiness for applications built on DNNs and thus hinders their further deployment in real world applications especially in the critical domains like healthcare \citep{qiu_large_2023}. Thus far, adversarial training (AT) has been the most effective defense against adversarial attacks \citep{athalye_obfuscated_2018}. 
AT is typically formulated as a min-max optimization problem:
\begin{equation}
    \arg \min_{\bm{\theta}} \mathbb{E}[\arg \max_{\bm{\delta}} \mathcal{L}(\bm{x}+\bm{\delta}; \bm{\theta})]
\end{equation}
where the inner maximization searches for the perturbation $\bm{\delta}$ to maximize the loss, while the outer minimization searches for the model parameters $\bm{\theta}$ to minimize the loss on the perturbed examples. 

One major issue of AT is that it is prone to overfitting \citep{rice_overfitting_2020, wong_fast_2020}. Unlike in standard training (ST), overfitting in AT, a.k.a. robust overfitting \citep{rice_overfitting_2020}, significantly impairs adversarial robustness. Many efforts \citep{li_understanding_2023, wu_adversarial_2020, dong_exploring_2022,liu_mitigating_2023,liu2023rethinking} have been made to understand robust overfitting and mitigate its effect. One promising solution is data augmentation (DA), which is a common technique to prevent ST from overfitting. However, many studies \citep{rice_overfitting_2020, wu_adversarial_2020, gowal_uncovering_2021, rebuffi_data_2021} have revealed that advanced DA methods, originally proposed for ST, often fail to improve adversarial robustness. 
Therefore, DA is usually combined with other regularization techniques such as Stochastic Weight Averaging (SWA) \citep{rebuffi_data_2021}, Consistency regularization \citep{tack_consistency_2022} and Separate Batch Normalization \citep{addepalli_efficient_2022} to improve its effectiveness.
However, recent work \citep{li_data_2023} demonstrated that DA alone can significantly improve AT if it has strong diversity and well-balanced hardness. This suggests that ST and AT may require different DA strategies, especially in terms of hardness. It is thus necessary to design DA schemes dedicated to AT.

\begin{figure*}
    \centering
    \includegraphics[width=0.9\linewidth, trim=0 180 0 145, clip]{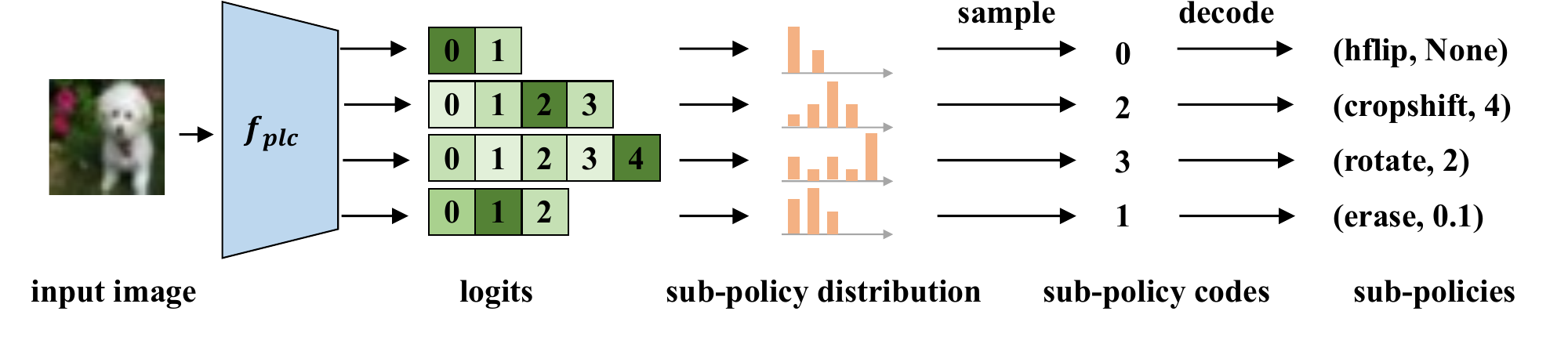}
    \caption{\textbf{An example of the proposed augmentation sampling procedure}. The policy model takes an image as input and outputs logit values 
    defining multiple, multinomial, probability distributions corresponding to different sub-policies. A sub-policy code is created by sampling from each of these distributions, and decoded into a sub-policy, i.e., a transformation and its magnitude. These transformations are applied, in sequence, to augment the image.}
    \label{fig: policy network samples augmentation}
\end{figure*}

IDBH \citep{li_data_2023} is the latest DA scheme specifically designed for AT. Despite its impressive robust performance, IDBH employs a heuristic search method to manually optimize the DA. 
This search process requires a complete AT for every sampled policy, which induces prohibitive computational cost and scales poorly to large datasets and models. 
Hence, when the computational budget is limited, the hyperparameters for IDBH might be found using a reduced search space\footnote{Search space refers to the collection of all possible data augmentation policies. Each policy consists of a set of a set of sub-policies, a data augmentation method associated with a magnitude, and a probability distribution for sampling each sub-policy to apply for data augmentation (see {\cref{fig: policy network samples augmentation}} for an illustration).} and by employing a smaller model, leading to compromised performance. 

\begin{figure*}
    \centering
    \includegraphics[width=.9\linewidth]{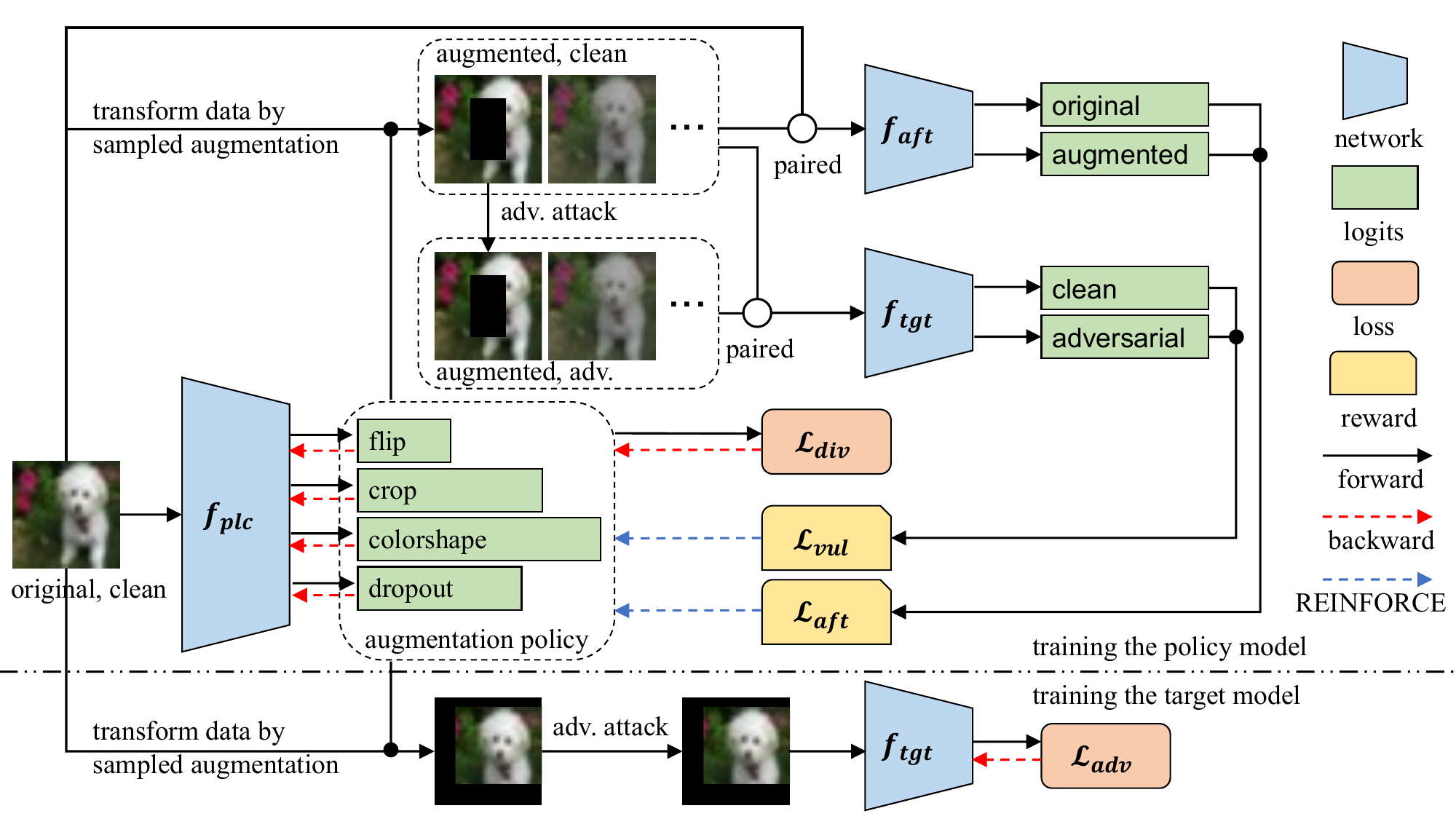}
    \caption{\textbf{An overview of the proposed method} (legend in the right column). The top part shows the pipeline for training the policy model, $f_{plc}$, while the bottom illustrates the pipeline for training the target model, $f_{tgt}$. $f_{aft}$ is a model pre-trained on clean data without any augmentation, which is used to measure the distribution shift caused by data augmentation. Please refer to \cref{sec: method} for a detailed explanation.}
    \label{fig: method overview}
\end{figure*}

Another issue is that IDBH, in common with other conventional DA methods such as AutoAugment \citep{cubuk_autoaugment_2019} and TrivialAugment \citep{muller2021trivialaugment}, applies the same strategy to all samples in the dataset throughout training. 
The distinctions between different training samples, and between the model checkpoints at different stages of training, are neglected. We hypothesize that different data samples at the same stage of training, as well as the same sample at the different stages of training, demand different DAs. Hence, we conjecture that an improvement in robustness could be realized by customizing DA for data samples and training stages.

To address the above issues, this work proposes a bi-level optimization framework (see \cref{fig: method overview}) to automatically learn \textbf{A}dversarial \textbf{R}obustness by \textbf{O}nline \textbf{I}nstance-wise \textbf{D}ata-augmentation (AROID).
To the best of our knowledge, \textbf{AROID is the first automated DA method specific to adversarial robustness}.
AROID employs a multi-head DNN-based policy model to map a data sample to a DA policy (see \cref{fig: policy network samples augmentation}). This DA policy is defined as a sequence of pre-defined transformations applied with strength determined by the output of the policy model. This policy model is optimized, alongside the training of the target model, towards three novel objectives to achieve a target level of hardness and diversity.  
DA policies, therefore, are customized for each data instance and evolve with the target network as training progresses. This in practice produces a more globally optimal DA policy and thus benefits robustness. Importantly, the proposed policy learning objectives, in contrast to the conventional ones like validation accuracy \citep{cubuk_autoaugment_2019}, do not reserve a subset of the training data for validation and do not rely on prohibitively expensive inner loops for training the target model to evaluate the rewards of the sampled policies. The former ensures the entire training set is available for training to avoid potential data scarcity. The latter enables policy optimization to be much more efficient and scalable so that it is more practical for AT. Compared to IDBH in particular, this allows our approach to explore a larger space of DAs. Taking an example of optimizing the DA for CIFAR10 and PRN18, \uline{AROID took 9 hours using an A100 GPU, IDBH took 412 hours using an A100 GPU, and AutoAugment took 5000 hours using a P100 GPU} \citep{hataya_faster_2020}.

Extensive experiments show that AROID outperforms all competitive DA methods across various datasets and model architectures while being more efficient than the previous best method (IDBH). \textbf{AROID achieves state-of-the-art robustness for DA methods} on the standard benchmarks. Besides, AROID outperforms, regarding accuracy and robustness, state-of-the-art AT methods. 
It also complements such robust training methods and can be combined with them to improve robustness further. 

\section{Related Work}

\textbf{Robust training}. To mitigate overfitting in AT, many methods other than DA, have been previously proposed. One line of works, IGR \citep{ross_improving_2018}, CURE \citep{moosavi2019robustness}, AdvLC \citep{li_understanding_2023}, discovered a connection between adversarial vulnerability and the smoothness of input loss landscape, and promoted robustness by smoothing the input loss landscape. Meanwhile, \cite{wu_adversarial_2020} and \cite{chen_robust_2021} found that robust generalization can be improved by a flat weight loss landscape and proposed AWP and SWA, respectively, to smooth the weight loss landscape during AT. RWP \citep{yu_robust_2022} and SEAT \citep{wang_self-ensemble_2022} were later proposed to further refine AWP and SWA, respectively, to increase robustness. 
SCARL \citep{kuang2023semantically} incorporated semantic information into adversarial training.
IBD \citep{kuang_improving_2023} distilled prior knowledge from a robust pre-trained model to enhance adversarial robustness.
Many works, including MART \citep{wang_improving_2020}, LAS-AT \citep{jia2022adversarial}, ISEAT \citep{li_improved_2023}, considered the difference between individual training instances and improved AT through regularizing in an instance-wise manner. Our proposed approach is also instance-wise, but contrary to existing methods 
tackles robust overfitting via DA instead of robust regularization. As shown in \cref{sec: rt robustness}, it works well alone and, more importantly, complements the above techniques.

\textbf{Data augmentation for ST}.
Although DA has been a common practice in many fields, we only review vision-based DA in this section as it is most related to our work. In computer vision, DA can be generally categorized as: basic, composite and mixup. Basic augmentations refer to a series of image transformations that can be applied independently. They mainly include crop-based (Random Crop \citep{he_deep_2016}, Cropshift \citep{li_data_2023}, etc.), color-based (Brightness, Contrast, etc.), geometric-based (Rotation, Shear, etc.) and dropout-based (Cutout \citep{devries_improved_2017}, Random Erasing \citep{zhong_random_2020}, etc.) transformations. Composite augmentations denote the composition of basic augmentations. Augmentations are composed into a single policy/schedule usually through two ways: interpolation \citep{hendrycks_augmix_2020, wang_augmax_2021} and sequencing \citep{cubuk_autoaugment_2019, cubuk_randaugment_2020, muller2021trivialaugment}. MixUp \citep{zhang2017mixup}, and analogous works like CutMix \citep{yun2019cutmix}, can be considered as a special case of interpolation-based composition, which combines a pair of different images, instead of augmentations, as well as their labels to create a new image and its label. 

Composite augmentations by design have many hyperparameters to optimize. Most previous works, as well as the pioneering AutoAugment \citep{cubuk_autoaugment_2019}, tackled this issue using automated machine learning (AutoML). DA policies were optimized towards maximizing validation accuracy \citep{cubuk_autoaugment_2019, lin_online_2019, li_differentiable_2020, liu2021direct}, maximizing training loss \citep{zhang_adversarial_2020} or matching the distribution density between the original and augmented data \citep{lim_fast_2019, hataya_faster_2020}. Optimization here is particularly challenging since DA operations are usually non-differentiable. Major solutions seek to estimate the gradient of DA learning objective w.r.t. the policy generator or DA operations using, e.g., policy gradient methods \citep{cubuk_autoaugment_2019, zhang_adversarial_2020, lin_online_2019} or reparameterization trick \citep{li_differentiable_2020, hataya_faster_2020}. Alternative optimization techniques include Bayesian optimization \citep{lim_fast_2019} and population-based training \citep{ho_population_2019}. Noticeably, several works like RandAugment \citep{cubuk_randaugment_2020} and TrivialAugment \citep{muller2021trivialaugment} found that if the augmentation space and schedule were appropriately designed, competitive results could be achieved using a simple hyperparameter grid search or fixed hyperparameters. 
This implies that in ST these advanced yet complicated methods may not be necessary. However, it remains an open question if simple search can still match these advanced optimization methods in AT. 
Besides, instance-wise DA strategy was also explored in \cite{cheung_adaaug_2022,miao_learning_2023} for ST.
Our method is the first automated DA approach specific for AT. We follow the line of policy gradient methods to enable learning DA policies. A key distinction here is that our policy learning objective is designed to guide the learning of DA policies towards improved robustness for AT, while the objective of the above methods is to increase accuracy for ST.

\section{Method} \label{sec: method}
We propose a method to automatically learn DA alongside AT to improve robust generalization. An \textbf{instance-wise} DA policy is produced by a policy model and learned by optimizing the policy model towards three novel objectives. Updating of the policy model and the target model (the one being adversarially trained for the target task) alternates throughout training (the policy model is updated every $K$ updates of the target model), yielding an \textbf{online} DA strategy. This online, instance-adaptive, strategy produces different augmentations for different data instances at different stages of training.


The following notation is used. $\bm{x} \in \mathbb{R}^d$ is a $d$-dimensional sample whose ground truth label is $y$. $\bm{x}_i$ refers to $i$-th sample in a dataset. 
The model is parameterized by $\bm{\theta}$. $\mathcal{L}(\bm{x}, y; \bm{\theta})$ or $\mathcal{L}(\bm{x}; \bm{\theta})$ for short denotes the predictive loss evaluated with $\bm{x}$ w.r.t. the model $\bm{\theta}$ (Cross-Entropy loss was used in all experiments). $\rho(\bm{x}; \bm{\theta})$ computes the adversarial example of $\bm{x}$ w.r.t. the model $\bm{\theta}$. $p_i(\bm{x}; \bm{\theta})$ or $p_i$ for short refers to the output of the Softmax function applied to the final layer of the model, i.e., the probability at $i$-th logit given the input $\bm{x}$. 

\subsection{Modeling the DA Policy}
\label{sec: da policy modeling}

Following the design of IDBH \citep{li_data_2023} and TrivialAugment \citep{muller2021trivialaugment}, DA is implemented using four types of transformations: flip, crop, color/shape and dropout applied in order. We implement flip using HorizontalFlip, crop using Cropshift \citep{li_data_2023}, dropout using Erasing\footnote{Different from the original version applied at half chance, here erasing is always applied but the location and aspect ratio are randomly sampled from the given range.} \citep{zhong_random_2020}, and color/shape using a set of operations including Color, Sharpness, Brightness, Contrast, Autocontrast, Equalize, Shear (X and Y), Rotate, Translate (X and Y), Solarize and Posterize. A dummy operation, Identity, is included in each augmentation group to allow data to pass through unchanged. More details including the complete augmentation space are described in \cref{app: augmentation space}.

To customize the DA applied to each data instance individually, a policy model parameterized by $\bm{\theta}_{plc}$, is used to produce a DA policy conditioned on the input data (see \cref{fig: policy network samples augmentation}). The policy model employs a DNN backbone to extract features from the data, and multiple, parallel, linear prediction heads on the top of the extracted features to predict the policy. The policy model used in this work has four heads corresponding to the four types of DA described above.
The output of a head is converted into a multinomial distribution where each logit represents a pre-defined sub-policy, i.e., an augmentation operation associated with a strength/magnitude (e.g. ShearX, 0.1). Different magnitudes of the same operation are represented by different logits, so that each has its own chance of being sampled. A particular sequence of sub-policies to apply to the input image are selected based on the probabilities encoded in the four heads of the policy network.


\subsection{Objectives for Learning the Data Augmentation Policy}
The policy model is trained using three novel objectives: (adversarial) Vulnerability, Affinity and Diversity. 
These objectives are designed to learn data augmentations with strong diversity and appropriate hardness: requirements that have been shown to be effective for adversarial training \citep{li_data_2023}.

\subsubsection{Motivation}
\label{sec: policy objectives motivation}

Intuitively, enhancing the diversity and hardness of data augmentation should help mitigate robust overfitting by increasing the complexity of the training data.
Specifically, enhanced diversity increases the number of distinct data augmentations applied during training and expands the effective training set size \citep{gontijo-lopes_tradeoffs_2021}. 
Increasing hardness raises the difficulty level of the augmented data for the model to learn (adversarially), thereby reducing (robust) overfitting. 
However, if the hardness exceeds the level that the training model can fit, accuracy and even robustness will decline, despite the reduction in robust overfitting.
Therefore, to maximize performance, hardness should be carefully adjusted to balance between reducing robust overfitting and improving overall performance.
The optimal level of hardness should therefore be tailored to different models and training settings.

Understanding what kind of data augmentation is effective for adversarial training is not the focus of the current work so we refer the reader to \citep{li_data_2023} for a formal quantitative definition of diversity and hardness, along with extensive experimental evidence supporting the above reasoning.

\subsubsection{Objectives}

Vulnerability measures the loss variation caused by adversarial perturbation on the augmented data w.r.t. the target model:
\begin{align}
    \mathcal{L}_{vul}(\bm{x}; \bm{\theta}_{plc}) &= \mathcal{L}(\rho (\bm{\hat{x}}; \bm{\theta}_{tgt}); \bm{\theta}_{tgt}) - \mathcal{L}(\bm{\hat{x}}; \bm{\theta}_{tgt}) \nonumber \\ \text{where}\ \bm{\hat{x}} &= \Phi(\bm{x}; S(\bm{\theta}_{plc}(\bm{x})))
    \label{equ: adversarial vulnerability}
\end{align}
$\Phi(\bm{x}; S(\bm{\theta}_{plc}(\bm{x})))$ augments $\bm{x}$ by $S(\bm{\theta}_{plc}(\bm{x}))$, the augmentations sampled from the output distribution of policy model conditioned on $\bm{x}$, so $\bm{\hat{x}}$ is the augmented data.
A larger Vulnerability indicates that $\bm{x}$ becomes more vulnerable to adversarial attack after DA. 
A common belief about the relationship between training data and robustness is that AT benefits from adversarially hard samples\footnote{``Adversarially hard samples'' refer to samples that are difficult to classify correctly after being adversarially perturbed. 
The difficulty, or hardness, generally increases with the adversarial vulnerability of the original sample and the strength of the adversarial attack.
From the perspective of attack strength, adversarially hard samples are those perturbed by stronger attacks.
The statement ``AT benefits from adversarially hard samples'' can, therefore, be understood more broadly as meaning that training with stronger attacks will lead to more effective adversarial training and thus higher robustness. 
For example, multi-step AT is generally considered more effective than single-step AT \citep{madry_towards_2018}.
From the perspective of adversarial vulnerability, adversarially hard samples are those with higher vulnerability to attacks. 
Hard data augmentation can make data more susceptible to attacks, thereby producing adversarially hard samples.
Empirical evidence \citep{li_data_2023} suggests that adversarial training benefits from increasing the hardness of data augmentation within an appropriate range, as this helps mitigate robust overfitting and enhance performance.
} \citep{madry_towards_2018,li_data_2023}. 
From a geometric perspective, maximizing Vulnerability encourages the policy model to project data into the previously less-robustified space. 

Nevertheless, the maximization of Vulnerability, if not constrained, would likely favor those augmentations producing samples far away from the original distribution. Training with such augmentations was observed to degrade accuracy and even robustness when accuracy is overly reduced \citep{li_data_2023}.
Therefore, Vulnerability should be maximized while the distribution shift caused by augmentation is constrained:
\begin{equation}
    arg\max_{\bm{\theta}_{plc}}\ \mathcal{L}_{vul}(\bm{x}; \bm{\theta}_{plc})\ \ \text{s.t.}\ ds(\bm{x}, \bm{\hat{x}}) \leq D
\label{equ: maximizing adv vulnerability constrained}
\end{equation}
where $ds(\cdot)$ measures the distribution shift between two samples and $D$ is a constant. 
Directly solving \cref{equ: maximizing adv vulnerability constrained} is intractable, so we convert it into an unconstrained optimization problem by adding a penalty on the distribution shift as:
\begin{equation}
    arg\max_{\bm{\theta}_{plc}}\ \mathcal{L}_{vul}(\bm{x}; \bm{\theta}_{plc}) - \lambda \cdot ds(\bm{x}, \bm{\hat{x}})
\label{equ: maximizing hardness}
\end{equation}
where $\lambda$ is a hyperparameter and a larger $\lambda$ corresponds to a tighter constraint on distribution shift, i.e., smaller $D$. 
Distribution shift is measured using a variant of the Affinity metric \citep{gontijo-lopes_tradeoffs_2021}:
\begin{equation}
    ds(\bm{x}, \bm{\hat{x}}) = \mathcal{L}_{aft}(\bm{x}; \bm{\theta}_{plc}) = \mathcal{L}(\bm{\hat{x}}; \bm{\theta}_{aft}) - \mathcal{L}(\bm{x}; \bm{\theta}_{aft})
    \label{equ: affinity}
\end{equation}
Affinity captures the loss variation caused by DA w.r.t. a model $\bm{\theta}_{aft}$ (called the affinity model): a model pre-trained on the original data (i.e., without any data augmentation). Affinity increases as the augmentation proposed by the policy network makes data harder for the affinity model to correctly classify. 
By substituting \cref{equ: affinity} into \cref{equ: maximizing hardness}, we obtain an adjustable Hardness objective:
\begin{equation}
    \mathcal{L}_{hrd}(\bm{x}; \bm{\theta}_{plc}) = \mathcal{L}_{vul}(\bm{x}; \bm{\theta}_{plc}) - \lambda \cdot \mathcal{L}_{aft}(\bm{x}; \bm{\theta}_{plc})
\label{equ: hardness measure}
\end{equation}
This encourages the DA produced by the policy model to be at a level of hardness defined by $\lambda$ (larger values of $\lambda$ corresponding to lower hardness). 
Ideally, $\lambda$ should be tuned to ensure the distribution shift caused by DA is sufficient to benefit robustness while not being so severe as to harm accuracy.

Last, we introduce a Diversity objective to promote diverse DA. Diversity enforces a relaxed uniform distribution prior over the logits of the policy model, i.e., the output augmentation distribution:
\begin{equation}
    \mathcal{L}_{div}^h (\bm{x}) = \frac{1}{C} [- \sum_i^{p_i^h < l} \log(p_i^h) + \sum_j^{p_j^h > u} \log(p_j^h)]
\label{equ: diversity loss}
\end{equation}
$C$ is the total count of logits violating either lower ($l$), or upper ($u$) limits and $h$ is the index of the prediction head. Intuitively speaking, the Diversity loss penalizes overly small and large probabilities, helping to constrain the distribution to lie in a pre-defined range $(l, u)$. As $l$ and $u$ approach the mean probability, the enforced prior becomes closer to a uniform distribution, which corresponds to a highly diverse DA policy. Diversity encourages the policy model to avoid the over-exploitation of certain augmentations and to explore other candidate augmentations. Note that Diversity is applied to the color/shape head in a hierarchical way: type-wise and strength-wise inside each type of augmentation.

Combining the above three objectives together, the policy model is trained to optimize:
\begin{equation}
    arg \min_{\bm{\theta}_{plc}} \ -\mathbb{E}_{i \in B} \mathcal{L}_{hrd}(\bm{x}_i) + \beta \cdot \mathbb{E}_{h \in H} \mathcal{L}_{div}^h (\bm{x}; \bm{\theta}_{plc})
\label{equ: overall policy learning objective}
\end{equation}
where $B$ is the batch size and $\beta$ trades-off hardness against diversity. $\mathcal{L}_{div}^h$ is calculated across instances in a batch, so no need for averaging over $B$ like $\mathcal{L}_{hrd}$. 

\subsubsection{Mechanism} 

The Vulnerability objective is computed using feedback on adversarial vulnerability, measured by the variation in loss caused by adversarial perturbations, from the target model. 
The policy model learns from this feedback to determine which types and magnitudes of data augmentation (DA) elevates the adversarial vulnerability of augmented data. This learning raises the likelihood of applying such augmentations to the training data, thereby resulting in increased hardness.
Meanwhile, the Affinity objective is employed to limit DA's hardness to a level that does not compromise performance. Additionally, the Diversity objective prevents the over-reliance on specific DA methods, promoting exploration across a diverse spectrum of augmentation techniques. Together, these three objectives dictate the appropriate DA for each training sample.

\subsection{Optimization}
The entire training is a bi-level optimization process (\cref{alg: training dual models}): the target and policy models are updated alternately. This online training strategy adapts the policy model to the varying demands for DA from the target model at the different stages of training. The target model is optimized using AT with the augmentation sampled from the policy model:
\begin{equation}
    arg \min_{\bm{\theta}_{tgt}} \mathcal{L}(\rho (\Phi(\bm{x}; S(\bm{\theta}_{plc}(\bm{x})));\bm{\theta}_{tgt}); \bm{\theta}_{tgt})
\end{equation}
After every $K$ updates of the target model, the policy model is updated using the gradients of the policy learning loss as follows:
\begin{align}
    \frac{\cref{equ: overall policy learning objective}}{\partial \bm{\theta}_{plc}} = - \frac{\partial \mathbb{E}_{i \in B} \mathcal{L}_{hrd}(\bm{x}_i)}{\partial \bm{\theta}_{plc}} + \beta \frac{\mathbb{E}_{h \in H} \mathcal{L}_{div}^h (\bm{x})}{\partial \bm{\theta}_{plc}}
\end{align}
The latter can be derived directly, while the former $\frac{\partial \mathcal{L}_{hrd}}{\partial \bm{\theta}_{plc}}$ cannot because the involved augmentation operations are non-differentiable. To estimate these gradients, we apply the REINFORCE algorithm \citep{williams_simple_1992} with baseline trick to reduce the variance of gradient estimation. It first samples $T$ augmentations, named trajectories, in parallel from the policy model and then computes the real Hardness value, $\mathcal{L}_{hrd}^{(t)}$, using \cref{equ: hardness measure} independently on each trajectory $t$. The gradients are estimated (see \cref{app: detailed derivation} for derivation) as follows: 
\begin{align}
    \frac{1}{B\cdot T}\sum_{i=1}^B\sum_{t=1}^T \sum_{h=1}^H \frac{\partial \log(p_{(t)}^h(\bm{x}_i))}{\partial \bm{\theta}_{plc}} [\mathcal{L}_{hrd}^{(t)}(\bm{x}_i) - \tilde{\mathcal{L}_{hrd}}]
\end{align}
$p_{(t)}^h$ is the probability of the sampled sub-policy at the $h$-th head and $\tilde{\mathcal{L}_{hrd}}=\frac{1}{T}\sum_{t=1}^T\mathcal{L}_{hrd}^{(t)}(\bm{x}_i)$ is the mean $\mathcal{L}_{hrd}$ (the baseline used in the baseline trick) averaged over the trajectories. \cref{alg: training policy model} illustrates one iteration of updating the policy model.
Note that, when one model is being updated, backpropagation is blocked through the other. The affinity model, used in calculating the Affinity metric, is fixed throughout training.

\begin{algorithm}[]
\SetAlgoCaptionLayout{small}
\SetAlgoCaptionSeparator{.}
\SetAlgoLined
\DontPrintSemicolon
\For{$i=1$ \KwTo $M$}{
    \tcp{for every $K$ iterations}
    \If{$i\ \% K == 0$}{
    \tcp{update the policy model by \cref{alg: training policy model}}
    }
    \tcp{the policy distribution}
    $d = \bm{\theta}_{plc}(\bm{x}_i)$ \;
    \tcp{sample \& apply augmentations}
    $\bm{\hat{x}}_i = \Phi(\bm{x}_i; S(d))$ \;
    $L = \mathcal{L}(\rho (\bm{\hat{x}}_i; \bm{\theta}_{tgt}); \bm{\theta}_{tgt})$ \;
    \tcp{update the target model}
    $\bm{\theta}_{tgt} = \bm{\theta}_{tgt} - \alpha_{tgt} \cdot \nabla_{\bm{\theta}_{tgt}}L$\;
}   
\caption{\textbf{High-level training procedures of the proposed method}. $\alpha$ is the learning rate. $M$ is the number of iterations.}
\label{alg: training dual models}
\end{algorithm}

\begin{algorithm}[]
\SetAlgoCaptionLayout{small}
\SetAlgoCaptionSeparator{.}
\SetAlgoLined
\DontPrintSemicolon
$d = \bm{\theta}_{plc}(\bm{x})$\; 
\tcp{same $\bm{x}$ used by all traj.}
\For{$t=1$ \KwTo $T$}{
    $\bm{\hat{x}}_{(t)} = \Phi(\bm{x}, S(d))$ \;
    $\mathcal{P}_{(t)} = \prod_{h=1}^H p_{(t)}^h$ \tcp{\small prob of traj $t$}
    $\mathcal{L}_{hrd}^{(t)}$ \tcp{computed by \cref{equ: hardness measure}}
}
$\tilde{\mathcal{L}_{hrd}} = \frac{1}{T} \sum_{t=1}^T \mathcal{L}_{hrd}^{(t)}$ \tcp{\small mean $\mathcal{L}_{hrd}^{(t)}$}
$L = \frac{1}{T} \sum_{t=1}^T \log(\mathcal{P}_{(t)})[\mathcal{L}_{hrd}^{(t)} - \tilde{\mathcal{L}_{hrd}}]$\;
$\mathcal{L}_{div}^{(h)}$ \tcp{computed using \cref{equ: diversity loss}}
$L = -L + \beta \frac{1}{H} \sum_{h=1}^H \mathcal{L}_{div}^{(h)}$ \;
$\bm{\theta}_{plc} = \bm{\theta}_{plc} - \alpha_{plc} \cdot \nabla_{\bm{\theta}_{plc}}L$\;
\caption{\textbf{Pseudo code of training the policy model for one iteration}. $\bm{x}$ is randomly sampled from the entire dataset.}
\label{alg: training policy model}
\end{algorithm}

\subsection{Modes of Application}
\label{sec: method mode}
AROID can be used in two modes: online and offline. 
In the online mode, the policy and target models are jointly trained so that the policy model has to be retrained every time a new target model is trained. 
This adapts the DA policy to the target model on-the-fly which improves effectiveness but adds the extra cost of policy learning to that of adversarial training.
In the offline mode, the training of policy and target models are separate phases. 
A policy model is trained in advance (using online AROID), a step that is analogous to the hyperparameter optimization of other DA methods.
This pre-trained policy model is then subsequently used to train a new target model.
Specifically, at each epoch of training the target network a policy network checkpoint, saved at the corresponding epoch when using online AROID, is used  to sample DA policies for training the target model. 
When AROID is deployed in this offline mode, we refer to it as AROID-T, as it involves the transfer of the policy model. The standard mode of application is online, which we refer to simply as AROID.

\subsection{Efficiency}
The efficiency of AROID is dependent on the mode.
The cost of AROID is composed of two parts: policy learning and DA sampling. Policy learning can be one-time expense if AROID is used in offline mode. DA sampling requires only one forward pass of the policy model, which can be negligible because the policy model can be much smaller than the target model without hurting the performance. Therefore, AROID in offline mode is roughly as efficient as other regular DA methods.

In online mode, in the worst case, AROID adds about 43.6\% extra computation to baseline AT (see calculation in \cref{app: efficiency policy learning}) when $T=8$ and $K=5$. This is less than the overhead 52.5\% of the state-of-the-art AT method LAS-AT \citep{jia2022adversarial} and substantially less than the search cost of IDBH and AutoAugment (compared in \cref{sec: efficiency}). Furthermore, we observed that AROID can still achieve robustness higher than other competitors with a much smaller policy model (\cref{app: ablation policy model}), reduced $T$ and increased $K$ (\cref{sec: efficiency}) for improved efficiency. For example, setting $T=4$ and $K=20$, the overhead is only about 10\% compared to baseline AT.

Another efficiency concern, as for all other deep learning methods, is hyperparameter optimization. 
We discuss below how this can be done efficiently so that AROID can be easily adapted to a new setting.
First, as shown in \cref{app: ablation hyperparameters}, most of our hyperparameters can transfer well among different training settings, so that only a light tuning is needed to achieve reasonably good performance for new setting. In most cases, only $\lambda$ needs to be tuned. 
Second, hyperparameter optimization can be accelerated by first searching with a cheap setting, such as $K=20$ and $T=4$, and then transferring the found values to the final setting, i.e., $K=5$ and $T=8$.
Note that our hyperparameter tuning process is not different from others.

\begin{table*}[tbp]
\centering
\caption{\textbf{The performance of various DA methods}. The \textbf{best} and \underline{second best} results are highlighted in each column. RandomCrop is the baseline DA consists of horizontal flip and random crop with 4 padding.}
\label{tab: data augmentation robustness}
\begin{tabular}{@{}lcccccccccc@{}}
\toprule
\multirow{3}{*}{DA Method} & \multicolumn{4}{c}{CIFAR10}                                                               & \multicolumn{4}{c}{CIFAR100}                                                  & \multicolumn{2}{c}{Imagenette}              \\ \cmidrule(l){2-11} 
                           & \multicolumn{2}{c}{WRN34-10}                & \multicolumn{2}{c}{ViT-B/4}                 & \multicolumn{2}{c}{WRN34-10}                & \multicolumn{2}{c}{PRN18}       & \multicolumn{2}{c}{ViT-B/16}                \\ \cmidrule(l){2-11} 
                           & Acc.                 & Rob.                 & Acc.                 & Rob.                 & Acc.                 & Rob.                 & Acc.           & Rob.           & Acc.                 & Rob.                 \\ \midrule
RandomCrop                   & 85.83                & 52.26                & 83.04                & 46.72                & 61.44                & 27.98                & 55.04          & 24.83          & 92.73                & 66.47                \\
Cutout                     & 86.95                & 52.89                & 83.61                & 48.67                & 59.04                & 27.51                & 57.37          & 24.51          & 93.27                & 67.20                \\
CutMix                     & 86.88                & 53.38                & 80.83                & 47.24                & 58.57                & 27.49                & 57.32          & 25.54          & 93.87                & {\ul 70.20}          \\
AutoAugment                & 87.71                & 54.60                & 81.96                & 47.47                & {\ul 64.10}          & {\ul 29.08}          & 58.51          & 25.28          & 95.13                & 67.60                \\
TrivialAugment             & 87.35                & 53.86                & 80.55                & 46.39                & 62.55                & 28.97                & 57.24          & 24.82          & {\textbf{95.25}} & 69.00                \\
IDBH                       & {\ul 88.61}          & {\ul 55.29}          & {\ul 85.09}          & {\ul 49.63}          & 60.93                & 29.03                & {\ul 59.38}          & {\ul 26.24}          & {\ul 95.20}          & 69.93                \\
AROID (ours)               & {\textbf{88.99}} & {\textbf{55.91}} & {\textbf{87.34}} & {\textbf{51.25}} & {\textbf{64.44}} & {\textbf{29.75}} & \textbf{60.17} & \textbf{26.56} & 94.88                & {\textbf{71.32}} \\ \bottomrule
\end{tabular}%
\end{table*}

\section{Experiments} \label{sec: experiments}
The experiments in this section were based on the following setup unless otherwise specified. 


\textbf{General set-ups.}
We used model architectures Vision Transformer (ViT-B/16 and ViT-B/4) \citep{dosovitskiy2020image}, WideResNet34-10 (WRN34-10) \citep{zagoruyko_wide_2016} and PreAct ResNet-18 (PRN18) \citep{he_identity_2016}. 
We evaluated on datasets CIFAR10/100 \citep{krizhevsky_learning_2009}, Imagenette\footnote{Imagenette is a subset of ImageNet consisting of 10 classes. We adopt a previous version (v1), \url{https://s3.amazonaws.com/fast-ai-imageclas/imagenette.tgz}, as suggested by \citet{mo_when_2022}.} and ImageNet \citep{deng_imagenet_2009}.

For CIFAR10/100, models were trained by stochastic gradient descent (SGD) for 200 epochs with an initial learning rate 0.1 divided by 10 at 50\% and 75\% of epochs. The momentum was 0.9, the weight decay was 5e-4 and the batch size was 128. 
The experiments on Imagenette and ImageNet followed a similar protocol as those on CIFAR10 except the following changes. 
For Imagenette, the weight decay was 1e-4, the total number of epochs was 40, and the learning rate was decayed at 36th and 38th epoch. 
The ViT-B/16 was pre-trained on ImageNet-1K. 
Gradient clipping was applied throughout training.
Note that CIFAR10 with ViT-B/4 is trained using the same setting as Imagenette with ViT-B/16.
For ImageNet, models were trained for 50 epochs with an initial learning rate 0.01 divided by 10 at 20th and 40th epoch. Models were pre-trained on ImageNet-1K. The weight decay was 0. 
Experiments were run on Nvidia Tesla V100 and A100. All results reported by us were averaged over 3 runs except for ImageNet due to the limit of computational resource.

\textbf{Adversarial set-ups.}
By default, we used $\ell_{\infty}$ PGD AT \citep{madry_towards_2018} with a perturbation budget, $\epsilon$, of 8/255. The number of steps was 10 and the step size was 2/255.
For ImageNet, the perturbation budget, $\epsilon$, was 4/255, the number of steps was 2 and the step size was $2\epsilon/3$. 
Following \cite{rice_overfitting_2020}, we tracked PGD10 robustness on the test set at the end of each epoch during training and selected the checkpoint with the highest PGD10 robustness, i.e., the "best" checkpoint to report robustness. 
Robustness was evaluated by AutoAttack \citep{croce_reliable_2020}.

\begin{table*}[]
\centering
\caption{\textbf{The performance of AROID-T, our method in offline mode.} Results compare the different settings of transferring pre-trained policy models with results obtained using AROID in online mode when  trained in the transfer destination setting.}
\label{tab: transfer result}
\begin{tabular}{@{}ccccccc@{}}
\toprule
\multirow{3}{*}{Policy source} &
  \multicolumn{2}{c}{CIFAR10$\rightarrow$CIFAR10} &
  \multicolumn{2}{c}{CIFAR10$\rightarrow$CIFAR10} &
  \multicolumn{2}{c}{CIFAR10$\rightarrow$CIFAR100} \\ \cmidrule(l){2-7} 
 &
  \multicolumn{2}{c}{WRN34-10$\rightarrow$WRN34-10} &
  \multicolumn{2}{c}{PRN18$\rightarrow$WRN34-10} &
  \multicolumn{2}{c}{WRN34-10$\rightarrow$WRN34-10} \\ \cmidrule(l){2-7} 
                            & Acc.           & Rob.           & Acc.           & Rob.           & Acc.           & Rob.           \\ \midrule
\multicolumn{1}{l}{AROID-T} & 88.76          & 55.61          & 86.17          & 50.70          & \textbf{64.97} & 29.67          \\
\multicolumn{1}{l}{AROID}   & \textbf{88.99} & \textbf{55.91} & \textbf{87.34} & \textbf{51.25} & 64.44          & \textbf{29.75} \\ \bottomrule
\end{tabular}
\end{table*}

\textbf{Configuration of AROID.}
Hyperparameters are optimized using grid search.
By default, $T=5$, $K=8$ and $\beta=0.8$ were used. 
The diversity limits $l$ and $u$ were 0.9 (0.8) 
\footnote{The value of $l$ and $u$ is a factor relative to the arithmetic mean chance, $\tilde{p}$, of sampling an augmentation in each group (prediction head), so the real absolute threshold value will be, e.g., $l \cdot \tilde{p}$. Taking an example of the Crop prediction head with 16 (1+15) magnitudes in total, $\tilde{p}=1 / 16$.}
and 4.0 respectively for CNNs (ViTs). 
$\lambda$ was 0.4-0.2-0.1 (decayed with the learning rate for better performance), 0.4 and 0.3 for WRN34-10, ViT-B/4 and PRN18 on CIFAR10, 0.3-0.1-0.01 and 0.2 for WRN34-10 and PRN18 on CIFAR100, and 0.3 for ViT-B/16 on Imagenette. 
The default backbone of the policy model was PRN18 except that ViT-B/16 (pre-trained on ImageNet-1K) was used for Imagenette\footnote{it was observed to be difficult for PRN18 to quickly fit Imagenette data to a reasonable degree in ST. Note that this ability is especially important when training on Imagenette because the total number of epochs (40) is much less than for the other datasets (200).}.

\cref{app: experiment setting} describes more implementation details of AROID and the competitive methods to be compared below.

\begin{table*}[]
\centering
\caption{\textbf{Evaluation of robust overfitting} for models trained with various data augmentation methods on CIFAR10/100 with WRN34-10. }
\label{tab: robust overfitting}
\begin{tabular}{@{}lllllllllllll@{}}
\toprule
\multicolumn{1}{c}{\multirow{3}{*}{DA Method}} &
  \multicolumn{6}{c}{CIFAR10} &
  \multicolumn{6}{c}{CIFAR100} \\ \cmidrule(l){2-13} 
\multicolumn{1}{c}{} &
  \multicolumn{3}{c}{Accuracy (\%)} &
  \multicolumn{3}{c}{Robustness (\%)} &
  \multicolumn{3}{c}{Accuracy (\%)} &
  \multicolumn{3}{c}{Robustness (\%)} \\ \cmidrule(l){2-13} 
\multicolumn{1}{c}{} &
  \multicolumn{1}{c}{Best} &
  \multicolumn{1}{c}{End} &
  Diff. &
  \multicolumn{1}{c}{Best} &
  \multicolumn{1}{c}{End} &
  Diff. &
  \multicolumn{1}{c}{Best} &
  \multicolumn{1}{c}{End} &
  Diff. &
  \multicolumn{1}{c}{Best} &
  \multicolumn{1}{c}{End} &
  Diff. \\ \midrule
baseline &
  85.8 &
  86.2 &
  \textbf{-0.3} &
  52.2 &
  46.6 &
  5.6 &
  61.4 &
  59.7 &
  1.7 &
  27.9 &
  24.2 &
  3.6 \\
Cutout &
  86.9 &
  87.4 &
  {\ul -0.4} &
  52.8 &
  51.0 &
  1.8 &
  59.0 &
  61.3 &
  -2.2 &
  27.5 &
  25.0 &
  2.4 \\
CutMix &
  86.8 &
  87.5 &
  -0.6 &
  53.3 &
  49.8 &
  3.5 &
  58.5 &
  62.8 &
  -4.3 &
  27.4 &
  26.2 &
  {\ul 1.2} \\
AutoAugment &
  87.7 &
  88.7 &
  -1.0 &
  54.6 &
  {\ul 54.0} &
  \textbf{0.5} &
  {\ul 64.1} &
  {\ul 64.6} &
  \textbf{-0.5} &
  {\ul 29.0} &
  27.1 &
  1.9 \\
TrivialAugment &
  87.3 &
  87.7 &
  {\ul -0.4} &
  53.8 &
  53.1 &
  {\ul 0.6} &
  62.5 &
  64.2 &
  -1.6 &
  28.9 &
  {\ul 27.3} &
  1.6 \\
IDBH &
  {\ul 88.6} &
  {\ul 88.9} &
  \textbf{-0.3} &
  {\ul 55.2} &
  53.4 &
  1.8 &
  60.9 &
  64.4 &
  -3.5 &
  29.0 &
  26.2 &
  2.8 \\
AROID (ours) &
  \textbf{88.9} &
  \textbf{89.2} &
  \textbf{-0.3} &
  \textbf{55.9} &
  \textbf{55.0} &
  0.9 &
  \textbf{64.4} &
  \textbf{65.9} &
  {\ul -1.5} &
  \textbf{29.7} &
  \textbf{28.9} &
  \textbf{0.8} \\ \bottomrule
\end{tabular}%
\end{table*}

\subsection{Benchmarking DA on Adversarial Robustness}
\label{sec: da robustness benchmark}

\cref{tab: data augmentation robustness} compares our proposed method against existing DA methods. \textbf{AROID outperforms all existing methods regarding robustness across all five tested settings}. The improvement over the previous best method is particularly significant for ViT-B on CIFAR10 (+1.62\%) and Imagenette (+1.12\%). Note that in most cases IDBH is the only method whose robustness is close to ours. However, our method is much more efficient than IDBH in terms of policy search (shown in \cref{sec: efficiency}). If our method is compared only to those methods with a computational cost the same or less than AROID's, i.e., excluding IDBH and AutoAugment, the improvement over the second best method is +2.05\%/2.58\%/0.78\%/1.12\%/1.02\% for the five experiments. Furthermore, we highlight the substantial improvement over the baseline of our method, +3.65\%/4.53\%/1.77\%/4.85\%/1.73\%, in these five settings.

In addition, \textbf{AROID also achieves the highest accuracy in four of the five tested settings}, and in the setting of Imagenette the accuracy gap between the best method and ours is marginal (0.37\%). Overall, \uline{our method significantly improves both accuracy and robustness, achieving a much better trade-off between accuracy and robustness}.
The consistent superior performance of our method, across various datasets (low and high resolution, simple and complex) and model architectures (CNNs and ViTs, small and large capacity), suggests that it has a good generalization ability.

\subsection{Offline vs. Online AROID}
\label{sec: transferrability policy}

This section evaluates the transferability of the learned policy models. It uses AROID in the offline mode (i.e. AROID-T as described in {\cref{sec: method mode}}), across three scenarios: 1) with the same dataset and model architecture; 2) across different datasets; 3) across different model architectures.
In scenario 1, a policy model is pre-trained on CIFAR10 for a WRN34-10 model and is applied to train a WRN34-10 model on CIFAR10. 
In scenario 2, a policy model is pre-trained on CIFAR10 for a WRN34-10 model and is applied to train a WRN34-10 model on CIFAR100. 
In scenario 3, a policy model is pre-trained on CIFAR10 for a PRN18 model and is applied to train a ViT-B/4 model on CIFAR10.

As shown in {\cref{tab: transfer result}}, AROID-T achieved accuracy and robustness comparable to its online counterpart, AROID. Importantly, AROID-T still outperforms previous data augmentation methods (\cref{tab: data augmentation robustness}) in terms of both accuracy and robustness. Notably, the cost of applying AROID-T is roughly the same as that of other data augmentation methods. Overall, these results demonstrate that AROID-T transfers well across various settings.

\subsection{Mitigating Robust Overfitting}

This section evaluates the effectiveness of our proposed method in mitigating robust overfitting.
Robust overfitting is measured, using the standard convention, as the difference between the best and end robustness. 
The results in \cref{tab: robust overfitting} demonstrate that compared to the baseline, AROID substantially reduces the degree of robust overfitting from 5.64\% to 0.91\% on CIFAR10 and from 3.69\% to 0.83\% on CIFAR100. AROID achieves the smallest robustness gap among all competitive methods on CIFAR100. 
Additionally, AROID achieves a robustness gap of 0.91\%, close to the minimum record of 0.52\% achieved by AutoAugment, while exhibiting significantly higher best and end robustness rates of +1.31\% and +0.92\%, respectively. 
Overall, these results suggest that \textbf{our method effectively mitigates robust overfitting.}

\subsection{Comparison of Policy Search Costs}\label{sec: efficiency}
We compare here the cost of policy search of AROID against other automated DA methods, i.e., AutoAugment and IDBH. Before comparison, it is important to be aware that the search cost for IDBH increases linearly with the size of search space, while the cost of AROID stays approximately constant. IDBH thus uses a reduced search space that is much smaller than the search space of AROID. However, reducing the search space depends on prior knowledge about the training datasets, which may not generalize to other datasets. Moreover, scaling IDBH to our larger search space is intractable, and it would be even more intractable if IDBH was applied to find DAs for each data instance at each stage of training, as is done by AROID.

Even in the most expensive configuration ($K=5$ and $T=8$), AROID is substantially cheaper than IDBH and AutoAugment regarding the cost of policy search as shown in \cref{tab: efficiency comparison}. The computational efficiency of AROID can be further increased by reducing the policy update frequency (increasing $K$) and/or decreasing the number of trajectories $T$,  while still matching the robustness of IDBH. If IDBH and AutoAugment were restricted to use the same, much lower, budget for searching for a DA policy, given the huge gap, we suspect that they may find nothing useful. 

\begin{table*}[]
\centering
\caption{\textbf{The cost of policy search for automated DA methods} using PRN18 on CIFAR10. AROID is used in online mode. The size of search space counts the possible combinations of probabilities and magnitudes. Our search space is uncountable due to its continuous range of probability, and is much larger than that of IDBH as it covers a much wider range of probabilities and magnitudes. Time denotes the total hours required for one search over the search space using an Nvidia A100 GPU for IDBH and AROID and a P100 GPU for AutoAugment (data is copied from \cite{hataya_faster_2020}).}
\label{tab: efficiency comparison}
\begin{tabular}{@{}lccccccccr@{}}
\toprule
\multirow{2}{*}{Method} &
  \multirow{2}{*}{K} &
  \multirow{2}{*}{T} &
  \multirow{2}{*}{Acc.} &
  \multirow{2}{*}{Rob.} &
  \multicolumn{4}{c}{Search Space} &
  \multirow{2}{*}{Time} \\ \cmidrule(lr){6-9}
\multicolumn{1}{c}{} &    &   &             &             & Prior & Probability & Magnitude & Size        & \multicolumn{1}{c}{} \\ \midrule
AutoAugment          & -  & - & 83.27    & 49.20        & No & discrete    & discrete  & $2.9\times10^{32}$          & 5000               \\
IDBH                 & -  & - & {\ul 84.23}    & 50.47   &  Yes     & discrete    & discrete  & 80          & 412.83               \\
AROID                & 5  & 8 & \textbf{84.68} & \textbf{50.57} & No & continuous  & discrete  & uncountable & 9.51                 \\
AROID               & 20 & 8 & 84.11          & 50.45         & No & continuous  & discrete  & uncountable & {\ul 6.85}           \\
AROID                & 20 & 4 & 83.63          & {\ul 50.52}  & No & continuous  & discrete  & uncountable & \textbf{6.24}        \\ \bottomrule
\end{tabular}%
\end{table*}

\subsection{Comparison with State-of-the-art Robust Training Methods} 
\label{sec: rt robustness}

\cref{tab: rt rob} compares our method against state-of-the-art robust training methods. It can be seen that AROID substantially improves vanilla AT in terms of  accuracy (by 3.16\%) and robustness (by 3.65\%). This improvement is sufficient to boost the performance of vanilla AT to surpass the state-of-the-art robust training methods like SEAT and LAS-AWP in terms of both accuracy and robustness. This suggests that our method achieved a better trade-off between accuracy and robustness while boosting robustness.

More importantly, our method, as it is based on DA, can be easily integrated into the pipeline of existing robust training methods and, as our results show, is complementary to them. 
Our method was combined with other AT methods in the same way as any other data augmentation method: simply by using the sampled data augmentation policy to augment the data before generating adversarial examples. 
The update of the policy model is independent of the training method used.
By combining with SWA and/or AWP, our method substantially improves robustness even further while still maintaining an accuracy higher than that achieved by others methods. 
It is worth noting that CutMix combined with SWA is widely recognized as a strong baseline for data augmentation. 
Our approach surpasses this baseline when combined with SWA as well.

\begin{table}[]
    \centering

    \caption{\textbf{The performance of various robust training (RT) methods with baseline and our augmentations} for WRN34-10 on CIFAR10.}   
    \label{tab: rt rob}
    
    \begin{tabular}{@{}lccc@{}}
    \toprule
    RT method & DA method & Acc. & Rob. \\ \midrule
    AT         & RandomCrop     & 85.83         & 52.26          \\
    AT-SWA     & RandomCrop     & 84.30             & 54.29              \\
    AT-AWP    & RandomCrop     & 85.93             & 54.34              \\
    AT-RWP    & RandomCrop     & 86.86          & 54.61          \\
    MART       & RandomCrop     & 84.17              & 51.10              \\
    MART-AWP   & RandomCrop     & 84.43              & 54.23              \\
    SEAT      & RandomCrop     & 86.44    & {\ul 55.67}    \\
    LAS-AT    & RandomCrop     & 86.23              & 53.58              \\
    LAS-AWP   & RandomCrop     & \textbf{87.74}              & 55.52              \\
     AT-SWA    & CutMix & \uline{87.65} & \textbf{56.03} \\ \midrule
     AT        & AROID (ours) & \textbf{88.99} & 55.91 \\
    AT-SWA     & AROID (ours) & 87.84          & 56.67          \\
    AT-AWP     & AROID (ours) & 87.94          & {\ul 56.98}          \\
    AT-AWP-SWA & AROID (ours) & {\ul88.39} & \textbf{57.03} \\ \bottomrule
    \end{tabular}%
\end{table}

\subsection{Generalization to Alternative AT Methods}

To further test the generalizability of AROID to alternative AT methods, we integrate AROID with two more superior AT methods: TRADES \citep{zhang_theoretically_2019} and SCORE \citep{pang2022robustness}. Results are shown in \cref{tab: trades score}. 
AROID achieves highest accuracy and robustness among all the tested DA methods with both advanced AT methods. 
Overall, these results together with those in \cref{sec: rt robustness}, show that AROID generalizes well to various AT methods (PGD, TRADES, SCORE, AWP, SWA).

\begin{table}[]
\centering
\caption{\textbf{Comparison of various DA methods when trained by alternative AT methods} like TRADES and SCORE for on CIFAR10 with PRN18. $\lambda$ is 0.6 for TRADES and 0.3 for SCORE. The other hyperparameters are configured by default as specified in \cref{app: experiment setting}.}
\label{tab: trades score}
\begin{tabular}{@{}lcccc@{}}
\toprule
\multirow{2}{*}{DA Method} & \multicolumn{2}{c}{TRADES}                  & \multicolumn{2}{c}{SCORE}                   \\ \cmidrule(l){2-5} 
               & Acc. & Rob.    & Acc. & Rob.    \\ \midrule
RandomCrop     & {\ul 83.01} & 49.10 & 80.15 & 48.88 \\
Cutout         & 81.74 & 48.98 & 82.02 & 50.08 \\
AutoAugment    & 80.76 & 48.64 & 81.68 & 49.93 \\
TrivialAugment & 80.91 & 48.04 & 80.39 & 49.24 \\
IDBH           & 82.24 & {\ul 50.86} & {\ul 82.35} & {\ul 50.97} \\
AROID (ours)                     & {\textbf{84.04}} & {\textbf{51.33}} & {\textbf{82.69}} & {\textbf{51.18}} \\ \bottomrule
\end{tabular}
\end{table}

\subsection{Combining with Extra Data}

\begin{table*}[]
\centering
\caption{\textbf{The performance of our methods when trained with extra data} on CIFAR10. ``Activation'' refers to the activation function in the model architecture. ``Batch'' denotes the batch size. The results of PORT and HAT are copied from RobustBench.}
\label{tab: extra data}
\begin{tabular}{@{}lccclcccc@{}}
\toprule
\multicolumn{1}{c}{Method} & AT & Activation & Extra Data & \multicolumn{1}{c}{DA} & Epochs & Batch & Acc. & Rob. \\ \midrule
baseline & PGD10 & ReLU & 0.5M Real     & RandomCrop & 200 & 128 & {\ul 88.78}    & 57.95          \\
PORT     & PGD10 & ReLU & 10M Synthetic & RandomCrop & 200 & 128 & 86.68          & {\ul 60.27}    \\
ours     & PGD10 & ReLU & 0.5M Real     & AROID      & 200 & 128 & \textbf{92.38} & \textbf{61.49} \\ \midrule
baseline & PGD10 & ReLU & 0.5M Real     & RandomCrop & 400 & 512 & 89.66          & 58.73          \\
HAT      & HAT   & SiLU & 0.5M Real     & RandomCrop & 400 & 512 & 91.47    & 62.83 \\
ours     & PGD10 & ReLU & 0.5M Real     & AROID      & 400 & 512 & \textbf{92.48} & 62.60    \\
 BDM      & PGD10 & ReLU & 50M Synthetic & RandomCrop & 400 & 512 & 92.06 & {\ul 63.39} \\
 ours     & PGD10 & ReLU & 50M Synthetic & AROID      & 400 & 512 & {\ul 92.28} & \textbf{63.56}    \\\bottomrule
\end{tabular}
\end{table*}

The leading methods on the robustness benchmark RobustBench \citep{croce_robustbench_2021} heavily use extra data to augment adversarial training. 
We incorporate AROID with extra real data following \citet{carmon_unlabeled_2019} and compare it against PORT \citep{sehwag_robust_2022} and HAT \citep{rade_reducing_2022} which are ranked, to date, first and second respectively in RobustBench for the model architecture WRN34-10.
As shown in \cref{tab: extra data}, our method significantly improves both accuracy and robustness over the baseline methods. 
Our method also surpasses PORT regarding both accuracy and robustness.
Our method, compared to HAT, achieves a comparable robustness and a clearly higher accuracy exhibiting a better trade-off between accuracy and robustness.
Note that HAT employs a more effective AT method, HAT, and a different activation function, SiLU, both of which are known to boost performance. 

Next, we test whether AROID can be applied to enhance the state-of-the-art method BDM \citep{wang_better_2023}, which utilizes 50M synthetic data samples.
As shown in \cref{tab: extra data}, AROID achieves a marginal improvement over this baseline in terms of accuracy and robustness, indicating that AROID remains effective even in data-rich settings. 
However, it is observed that the performance improvement provided by AROID diminishes when compared to results without the additional 50M data. 
This reduction occurs because the robust overfitting in the baseline is largely mitigated by the additional data, and since AROID enhances adversarial training by alleviating robust overfitting, the scope for further improvement by AROID is consequently reduced.

Although the benefit of data augmentation diminishes when a large amount of synthetic data is incorporated for training on CIFAR10, this approach may not be as effective on more complex datasets such as ImageNet. 
As observed in \citep{azizi_synthetic_2023}, increasing synthetic ImageNet data beyond a certain limit (around 1.2M synthetic images) degrades model performance in high-resolution settings (256x256 and 1024x1024 pixels), while it consistently provides benefits in low-resolution setting (64x64 pixels). 
This degradation at high resolutions may be due to greater bias in the model and/or lower quality in the generated images at higher resolutions.

\begin{table}[]
    \centering
    \caption{\textbf{The result of AROID on ImageNet} with ConvNeXt-T.}
    \label{tab: da imagenet}
    \begin{tabular}{@{}lcc@{}}
    \toprule
    \multicolumn{1}{c}{DA method} & \multicolumn{1}{c}{Accuracy} & \multicolumn{1}{c}{Robustness} \\ \midrule
    RandomCrop                   & {\ul 71.22}                    & 36.22                    \\
    AutoAugment                & 70.42                    & {\ul 37.80}                    \\
    AROID (ours)               & \textbf{71.62}                & \textbf{40.40}                \\ \bottomrule
    \end{tabular}%
\end{table}

\begin{table*}[tbp]
\centering
\caption{\textbf{Robustness evaluation against more adversarial attacks}. PGD uses 50 steps and 10 restarts. CW and JITTER use 100 steps. Note that the abnormally superior PGD robustness but worse against other attacks of CutMix suggest a false security caused by obfuscated gradients.}
\label{tab: more attacks}
\begin{tabular}{lcccccccccc}
\toprule
\multicolumn{1}{c}{\multirow{2}{*}{DA Methods}} &
  \multicolumn{5}{c}{CIFAR10+WRN34-10} &
  \multicolumn{5}{c}{Imagenette+ViT-B/16} \\ \cmidrule(l){2-11}
\multicolumn{1}{c}{} &
  Clean &
  AA &
  PGD &
  CW &
  JITTER &
  Clean &
  AA &
  PGD &
  CW &
  JITTER \\ \hline
RandomCrop &
  85.8 &
  52.2 &
  55.5 &
  54.2 &
  53.5 &
  92.7 &
  66.4 &
  68.1 &
  68.4 &
  68.8 \\
Cutout &
  86.9 &
  52.8 &
  55.3 &
  55.0 &
  54.6 &
  93.2 &
  67.2 &
  68.4 &
  68.6 &
  69.4 \\
CutMix &
  86.8 &
  53.3 &
  \textbf{60.1} &
  56.9 &
  56.4 &
  93.8 &
  {\ul 70.2} &
  {\textbf{73.1}} &
  {\ul 71.8} &
  {\ul 72.2} \\
AutoAugment &
  87.7 &
  54.6 &
  58.8 &
  56.3 &
  55.6 &
  95.1 &
  67.6 &
  68.9 &
  69.8 &
  70.6 \\
TrivialAugment &
  87.3 &
  53.8 &
  57.4 &
  55.2 &
  55.4 &
  {\textbf{95.2}} &
  69.0 &
  70.9 &
  70.6 &
  71.5 \\
IDBH &
  {\ul 88.6} &
  {\ul 55.2} &
  58.2 &
  {\ul 57.3} &
  {\ul 56.9} &
  {\ul 95.2} &
  69.9 &
  70.2 &
  70.8 &
  71.6 \\
AROID (ours) &
  {\textbf{88.9}} &
  {\textbf{55.9}} &
  {\ul 59.6} &
  {\textbf{58.1}} &
  {\textbf{57.6}} &
  94.8 &
  {\textbf{71.3}} &
  {\ul 71.8} &
  {\textbf{72.8}} &
  {\textbf{73.1}} \\ 
  \bottomrule
\end{tabular}
\end{table*}

\subsection{Generalization to ImageNet}
To further test the generalizability and scalability of our method to a large-scale dataset, we train AROID on ImageNet \citep{deng_imagenet_2009} with ConvNeXt-T \citep{liu_convnet_2022}. Some DA methods are missing in this comparison due to limited computational resources (explained in \cref{app: config of da methods}). As shown in \cref{tab: da imagenet}, AROID significantly improves robustness over the baseline by 4.18\% and AutoAugment by 2.6\%. It also achieves the highest accuracy among the tested methods. Overall, AROID is able to scale and generalize to ImageNet.

The AROID hyperparameters were set to $\lambda=0.7$, $\beta=2$, $(l, u) = (0.8, 4.0)$, $T=20$ and $K=4$. 
As we did not have sufficient computational resources to fully optimize these hyperparameters on ImageNet performance is likely to be suboptimal and falls-short of the state-of-the-art result \citep{singh_revisiting_2023}. 
It has been observed in \citep{singh_revisiting_2023} that adversarial training on ImageNet prefers heavy data augmentation that is composed of RandAugment \citep{cubuk_randaugment_2020}, CutMix, MixUp and Random Erasing.
DA operations like CutMix and MixUp are not included in our DA search space. 
Incorporating these operations into our search space is thus expected to boost the performance of our method on ImageNet. 
We leave the exploration of this enhancement to the future.



\subsection{Robustness Evaluation with More Attacks} \label{app: more attack evaluation}

To further ensure our robustness evaluation is reliable, we additionally evaluate AROID and other related works using three more adversarial attacks PGD \citep{madry_towards_2018}, CW \citep{carlini_towards_2017} and JITTER \citep{schwinn_exploring_2023}. From the results shown in \cref{tab: more attacks} it can be seen that AROID is consistently superior under various adversarial attacks.

\subsection{Performance on Common Corruption Datasets}

This section assesses the generalization capability of the proposed method under input data distribution shifts, known as Out-Of-Distribution (OOD) testing. Following \citet{kireev_effectiveness_2022}, we trained models on the CIFAR10 training set and evaluated them on CIFAR10-C \citep{hendrycks_benchmarking_2019}. CIFAR10-C is created by applying 15 types of common visual corruptions to the CIFAR10 test set, representing visual corruption shifts encountered in the wild.

In \citet{kireev_effectiveness_2022}, only clean accuracy was evaluated on CIFAR10-C, focusing on the efficacy of adversarial training in improving robustness against common corruptions. However, this study emphasizes adversarial robustness. A recent study suggested that adversarial robustness is highly vulnerable to input distribution shifts \citep{li_oodrobustbench_2024}. Therefore, we also evaluated adversarial robustness on CIFAR10-C by conducting AutoAttack on the CIFAR10-C data.

As shown in \cref{tab: common corruption}, our proposed method achieves the highest accuracy and robustness among all competitive data augmentation methods, indicating excellent OOD generalization ability for both clean and robust performance under common corruption distribution shifts.

\begin{table}[]
\centering
\caption{\textbf{The performance of various DA methods on the common corruption dataset CIFAR10-C for WRN34-10.} Models were trained on CIFAR10 training set. }
\label{tab: common corruption}
\begin{tabular}{@{}lcccc@{}}
\toprule
\multicolumn{1}{c}{\multirow{2}{*}{Method}} & \multicolumn{2}{c}{CIFAR10} & \multicolumn{2}{c}{CIFAR10-C} \\ \cmidrule(l){2-5} 
\multicolumn{1}{c}{} & Acc. & Rob. & Acc. & Rob. \\ \midrule
RandomCrop & 85.83 & 52.26 & 76.70 & 36.69 \\
Cutout & 86.95 & 52.89 & 76.46 & 35.97 \\
CutMix & 86.88 & 53.38 & 77.48 & 36.91 \\
AutoAugment & 87.71 & 54.60 & 78.30 & 36.68 \\
TrivialAugment & 87.35 & 53.86 & 78.42 & 37.99 \\
IDBH & {\ul 88.61} & {\ul 55.29} & {\ul 79.37} & {\ul 38.15} \\
AROID (ours) & \textbf{88.99} & \textbf{55.91} & \textbf{80.61} & \textbf{39.72} \\ \bottomrule
\end{tabular}%
\end{table}

\begin{table*}[]
\centering
\caption{\textbf{The performance of baseline RandomCrop with larger models} on CIFAR10.}
\label{tab: efficiency model size comparison}
\begin{tabular}{@{}lcccccccc@{}}
\toprule
\multicolumn{1}{c}{\multirow{2}{*}{Data Augmentation}} &
  \multirow{2}{*}{Model} &
  \multirow{2}{*}{Model Size (M)} &
  \multicolumn{3}{c}{Accuracy(\%)} &
  \multicolumn{3}{c}{Robustness(\%)} \\ \cmidrule(l){4-9} 
\multicolumn{1}{c}{} &          &      & Best        & End         & Diff.       & Best        & End         & Diff.      \\ \midrule
AROID &
  WRN34-10 &
  46.2 &
  \textbf{88.99} &
  \textbf{89.29} &
  \textbf{-0.30} &
  \textbf{55.91} &
  \textbf{55.00} &
  \textbf{0.91} \\
RandomCrop           & WRN34-10 & 46.2 & 85.83       & 86.21       & {\ul -0.38} & 52.26       & 46.63       & 5.63       \\
RandomCrop           & WRN34-12 & 66.5 & {\ul 86.65} & {\ul 86.45} & 0.20        & 52.46       & {\ul 48.34} & {\ul 4.12} \\
RandomCrop           & WRN46-10 & 65.5 & 86.61       & 86.38       & 0.23        & {\ul 52.98} & 47.63       & 5.35       \\ \bottomrule
\end{tabular}
\end{table*}

\begin{table*}[]
\centering
\caption{\textbf{The performance of AROID with the original and the enlarged (with CutMix added) data augmentation space with and without SWA} for WRN34-10 on CIFAR10. Models were trained for 400 epochs, in contrast to 200 epochs used in \cref{tab: robust overfitting}, to better demonstrate the effect of adding CutMix in reducing robust overfitting.}
\label{tab: enlarged da space}
\begin{tabular}{@{}clllllll@{}}
\toprule
\multicolumn{1}{c}{\multirow{2}{*}{AT Method}} &
  \multicolumn{1}{c}{\multirow{2}{*}{DA Space}} &
  \multicolumn{3}{c}{Accuracy (\%)} &
  \multicolumn{3}{c}{Robustness (\%)} \\ \cmidrule(l){3-8} 
\multicolumn{1}{c}{} &
  \multicolumn{1}{c}{} &
  \multicolumn{1}{c}{Best} &
  \multicolumn{1}{c}{End} &
  \multicolumn{1}{c}{Diff.} &
  \multicolumn{1}{c}{Best} &
  \multicolumn{1}{c}{End} &
  \multicolumn{1}{c}{Diff.} \\ \midrule
\multirow{2}{*}{AT}  & Original        & \textbf{89.50} & \textbf{89.59} & \textbf{-0.09} & 55.56          & 53.33          & 2.23          \\
                     & Original+CutMix & 88.93          & 89.46          & -0.53          & \textbf{56.44} & \textbf{55.83} & \textbf{0.62} \\ \midrule
 \multirow{2}{*}{AT-SWA} & Original        & 88.71          & \textbf{90.40} & -1.69          & 57.31          & 55.05          & 2.26          \\
                     & Original+CutMix & \textbf{89.52} & 90.02          & \textbf{-0.50} & \textbf{57.32} & \textbf{57.14} & \textbf{0.18} \\ \bottomrule
\end{tabular}
\end{table*}

\subsection{Data Scaling vs. Model Scaling}

This section compares the effectiveness of scaling up data (our method) versus scaling up the model in enhancing adversarial training.
To test this, we trained AROID using the WRN34-10 model architecture (depth of 34 and widening factor of 10) and compared it to WRN34-12 and WRN46-10 architectures trained with RandomCrop DA. 
WRN34-12 and WRN46-10 were chosen because they have approximately 44\% and 42\% more parameters, respectively, than WRN34-10, which is comparable to the worst-case extra computational overhead, 43.6\%, caused by AROID.

As shown in \cref{tab: efficiency model size comparison}, AROID with WRN34-10 achieved the highest accuracy and robustness, greatly outperforming RandomCrop even when larger models were used. 
This suggests that \textbf{optimizing data augmentation, when implemented correctly, can be more effective than merely scaling up the model to boost performance}. 
The issue with RandomCrop and larger models is that, as indicated by the large gap between best and end robustness, scaling up models cannot effectively mitigate robust overfitting, resulting in poor generalization of robustness.

\subsection{Enlarging Policy Search Space}

\begin{figure*}
    \centering
    
    \subfloat[]{\includegraphics[width=.24\linewidth]{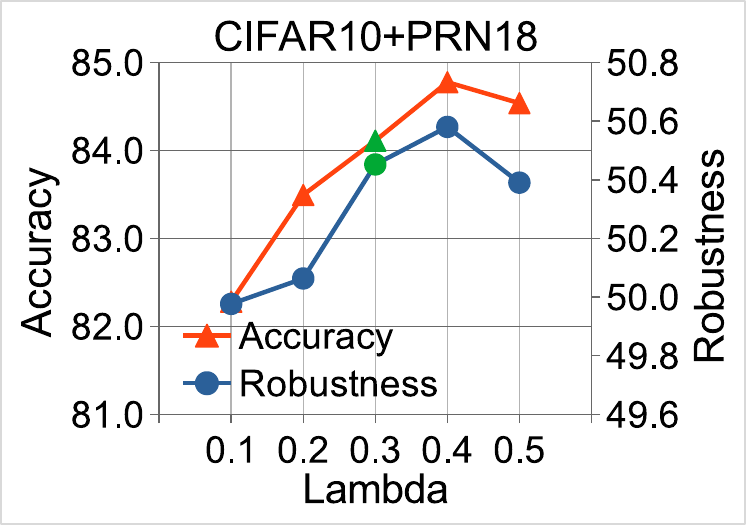} \label{fig: prn lambda}}
    \subfloat[]{\includegraphics[width=.24\linewidth]{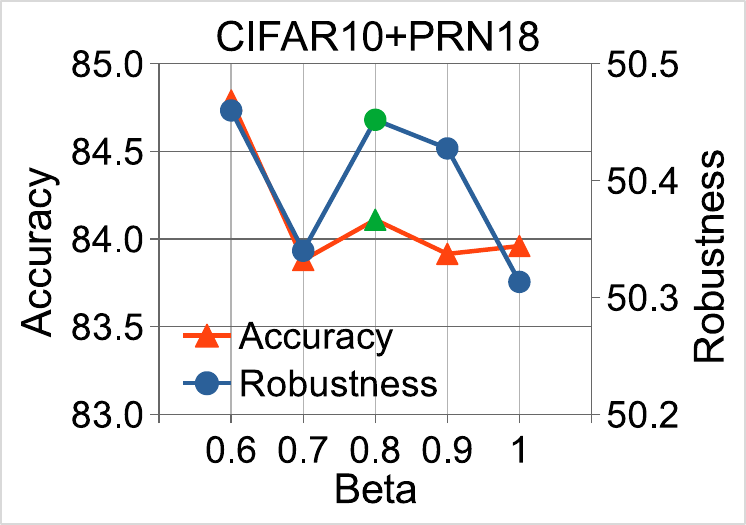} \label{fig: prn beta}}
    \subfloat[]{\includegraphics[width=.24\linewidth]{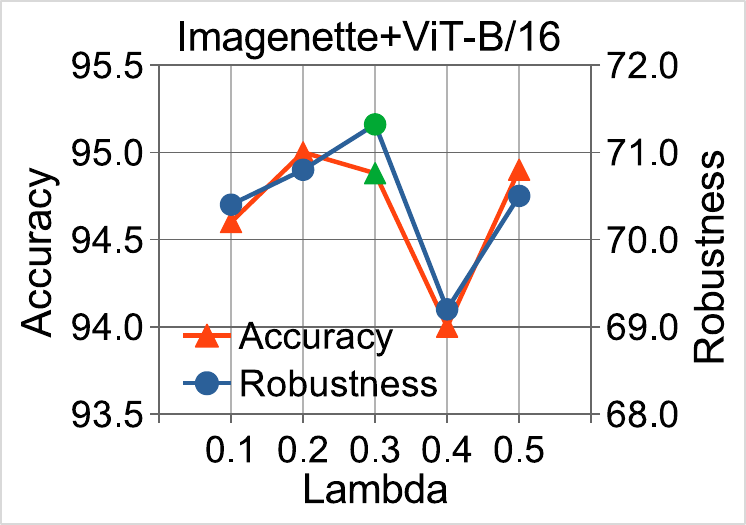} \label{fig:vit lambda}}
    \subfloat[]{\includegraphics[width=.24\linewidth]{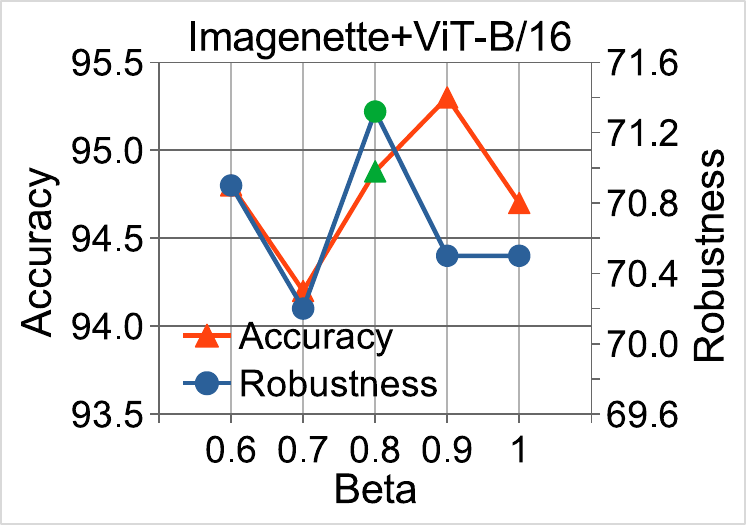} \label{fig: vit beta}}
    
    \subfloat[]{\includegraphics[width=.24\linewidth]{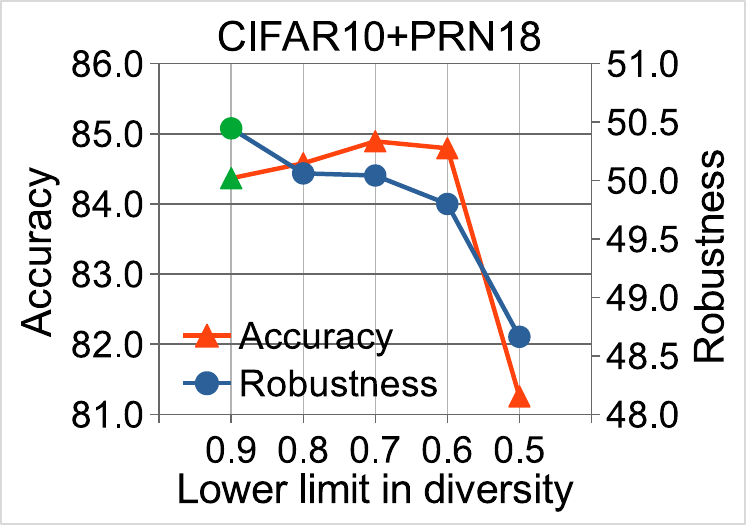} \label{fig: prn lower}}
    \subfloat[]{\includegraphics[width=.24\linewidth]{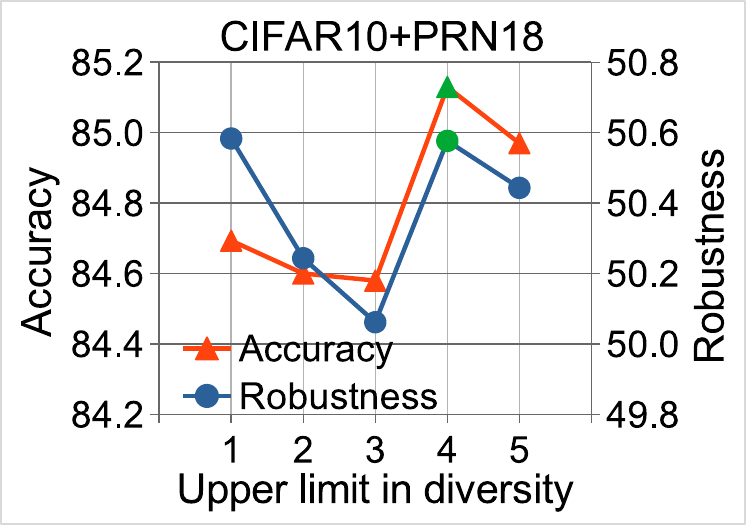} \label{fig: prn upper}}
    \subfloat[]{\includegraphics[width=.24\linewidth]{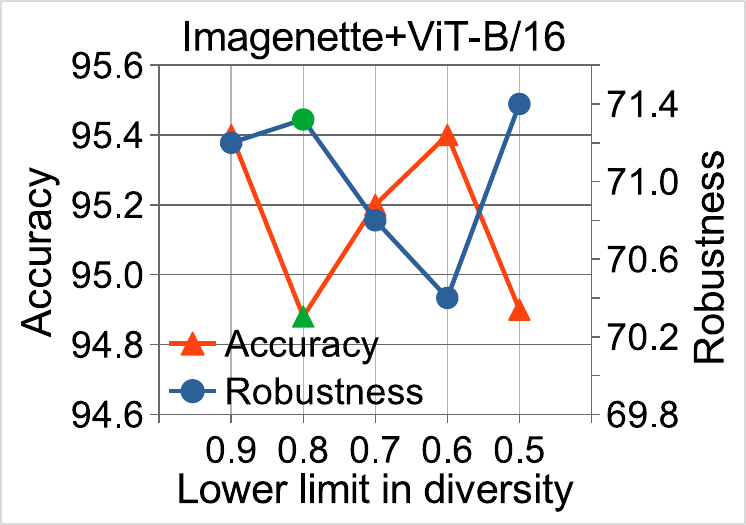} \label{fig: vit lower limit}}
    \subfloat[]{\includegraphics[width=.24\linewidth]{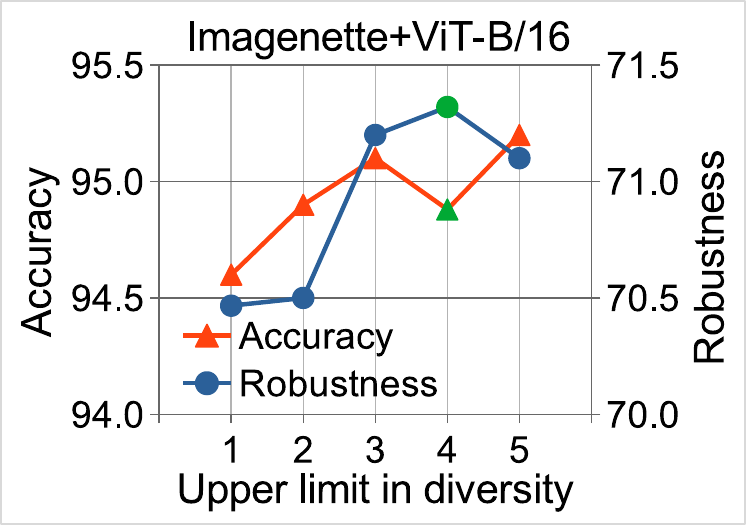} \label{fig: vit upper}}
    
    \subfloat[]{\includegraphics[width=.24\linewidth]{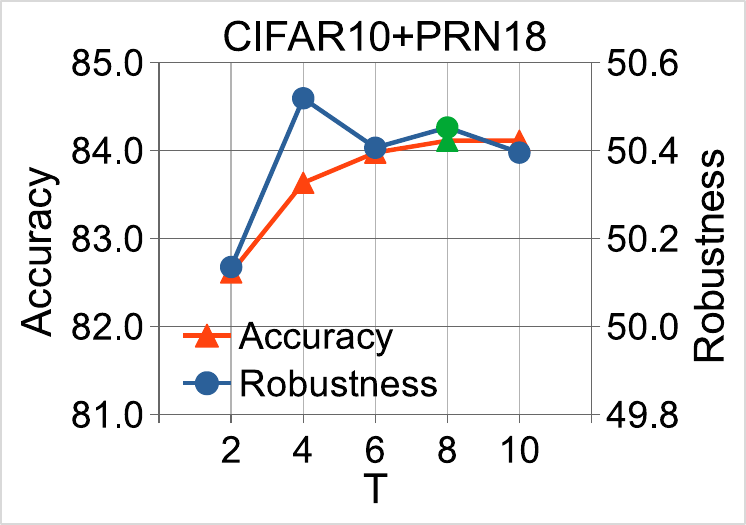} \label{fig: prn T}}
    \subfloat[]{\includegraphics[width=.24\linewidth]{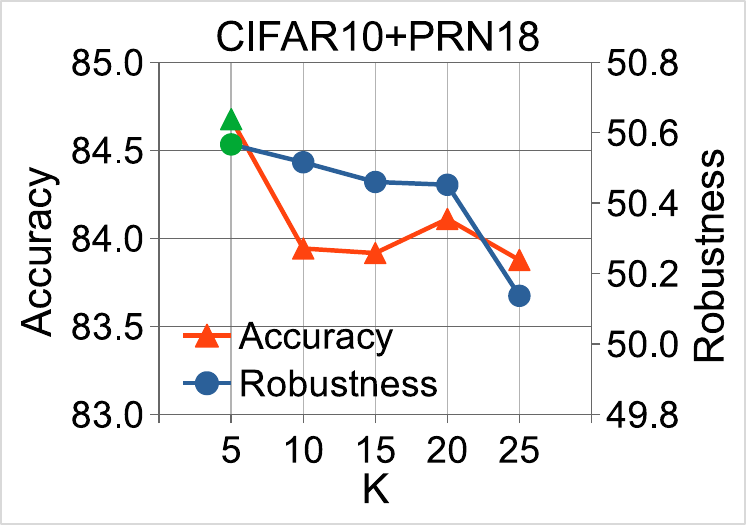} \label{fig: prn K}}
    \subfloat[]{\includegraphics[width=.24\linewidth]{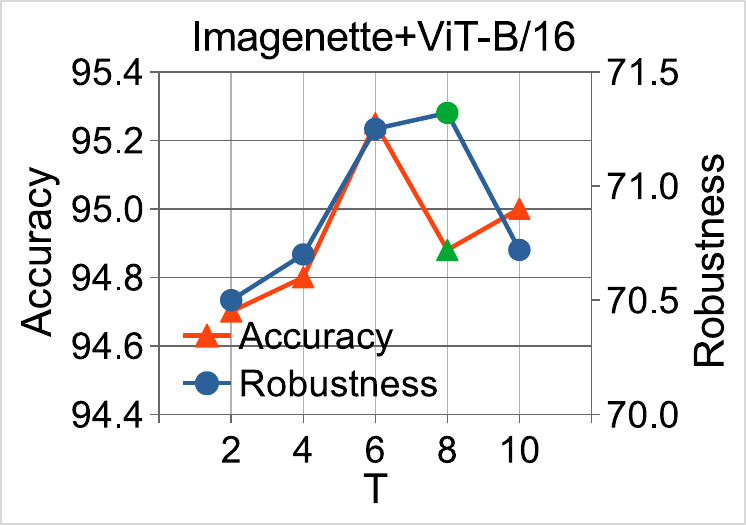} \label{fig: vit T}}
    \subfloat[]{\includegraphics[width=.24\linewidth]{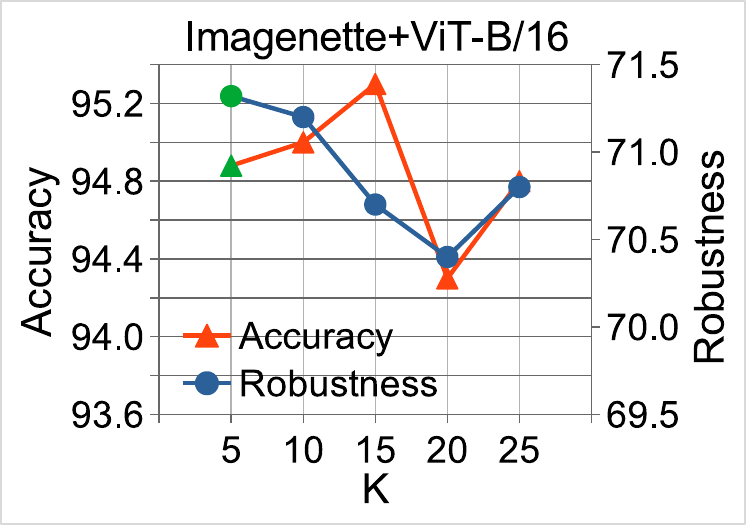} \label{fig: vit K}}   

    \caption{\textbf{Ablation study of hyper-parameters} $\lambda$, $\beta$, $l$, $u$, $T$ and $K$ for CIFAR10 with PRN18 (even rows) and Imagenette with ViT-B/16 (odd rows). The selected value for each hyper-parameter is marked green color.}
    \label{fig: ablation study}
\end{figure*}

This section assesses if enlarging policy search space can enhance AROID.
We conducted tests by adding CutMix to our policy search space as an additional transformation to be sampled and applied after the dropout transformation (please refer to {\cref{sec: da policy modeling}} for the specification of data augmentation policy structure). 
CutMix was chosen due to its effectiveness in adversarial training when combined with SWA \citep{rebuffi_data_2021}.

As shown in \cref{tab: enlarged da space}, \textbf{the inclusion of CutMix, compared to the original data augmentation space, results in reduced robust overfitting and improved best and end robustness, regardless of whether it is combined with SWA or not}. 
Additionally, incorporating CutMix even leads to a boost in best accuracy when combined with SWA. 
One possible account for this improvement is that the addition of CutMix increases the diversity of data augmentation in the learned policy, thereby mitigating robust overfitting and enhancing robust generalization (the reasons why diverse data augmentation mitigates robust overfitting are explained in \cref{sec: policy objectives motivation}).

However, it is important to note that not all data augmentation methods yield such benefits. 
The impact of incorporating additional data augmentation methods into the policy search space is specific to the nature of the augmentation techniques themselves. 
Toxic data augmentation methods, as observed in \citep{cubuk_randaugment_2020}, may not enhance, and in some cases, may even impair the performance of AROID if added to the search space. 
Overall, AROID can indeed benefit from an enlarged search space if implemented appropriately.

\subsection{Ablation Study} \label{app: ablation study}
This section verifies the sensitivity of our method to its hyperparameters and several design choices. 
The experiments were conducted on CIFAR10 with PRN18 and Imagenette with ViT-B/16. The default values of hyperparameters are the ones marked in green in \cref{fig: ablation study}.



\subsubsection{Hyperparameters}
\label{app: ablation hyperparameters}

\textbf{Policy update frequency $K$}. 
\cref{fig: prn K,fig: vit K} show that the highest accuracy and robustness were achieved when $K=5$, i.e., the lowest frequency under the test. This implies that AT benefits from a more ``up-to-date'' DA. Furthermore, it seems possible to trade accuracy for efficiency by choosing a larger value of $K$ (up to 20) while maintaining similarly high robustness. In general, the accuracy and robustness of our method declines with lower policy update frequency.

\textbf{Number of trajectories $T$}. 
\cref{fig: prn T,fig: vit T} show that high accuracy and robustness are achieved around $T=8$. This suggests that (1) there is a minimum requirement on the amount of trajectories for our policy gradient estimator to be accurate and, (2) our method may not benefit from increasing $T$ beyond 8.

\textbf{Strength of Affinity $\lambda$}.
As shown in \cref{fig: prn lambda,fig:vit lambda}, robustness first increases and then decreases within the tested range of value. This is consistent with the prior that AT benefits from appropriate hardness but degrade if data augmentations are overly hard \citep{li_data_2023}. 

\textbf{Strength of Diversity $\beta$}.
The performance within the tested range of value is close in \cref{fig: prn beta,fig: vit beta}, suggesting that the performance of AROID is not sensitive to the value of $\beta$. 
Nevertheless, this does not imply that Diversity is unnecessary in our policy learning.
On the contrary, it plays an important role in policy learning as shown in {\cref{sec: policy learning objective ablation study}}.

\textbf{Summary}.
We observe that, within the tested value range, hyper-parameters like $\lambda$, $\beta$, $T$ and $K$ have a quite similar trend in both settings, while the lower limit $l$ (\cref{fig: prn lower,fig: vit lower limit}) and upper limit $u$ (\cref{fig: prn upper,fig: vit upper}) in the diversity objective shows slightly different trends between the two settings. Despite the slightly different behaviors of a few hyper-parameters, the optimal value of hyper-parameters is observed to transfer across these two settings, i.e., they achieve reasonably good performance with a similar set of hyper-parameter values $T=8$, $K=5$, $l=0.8/0.9$, $u=4$, $\lambda=0.3$, $\beta=0.8$. We also find this setting transfers well across different AT methods of PGD, SCORE and TRADES since we can only tune the value of $\lambda$ while keep the rest unchanged to achieve reasonably good performance and outperform the other compared data augmentations.

\begin{table}[]
\centering
\caption{\textbf{The impact of removing each policy learning objective on the performance of AROID} for PRN18 and CIFAR10. The performance drop is marked by {\color{red}red} color.}
\label{tab: objectives ablation}
\begin{tabular}{@{}lcc@{}}
\toprule
\multicolumn{1}{c}{Policy Objectives} & Accuracy (\%)                        & Robustness (\%) \\ \midrule
AROID                      & \textbf{84.68}                       & \textbf{50.57}  \\
- Vulnerability            & 83.50 ({\color{red}-1.18}) & 49.95 ({\color{red}-0.62})   \\
- Affinity                 & 82.03 ({\color{red}-2.65})                        & 49.41 ({\color{red}-1.16})   \\
- Diversity                & 73.88 ({\color{red}-10.8})                        & 22.47 ({\color{red}-28.1})   \\ \bottomrule
\end{tabular}
\end{table}

\subsubsection{Policy Learning Objectives}
\label{sec: policy learning objective ablation study}

This section conducts an ablation study to evaluate the effect of each proposed policy learning objective on the performance of AROID.
As shown in {\cref{tab: objectives ablation}}, \textbf{removing any single policy learning objective leads to a considerable drop in both accuracy and robustness, indicating that each objective is crucial for learning an effective data augmentation policy}. 
Particularly, we observed that when Diversity is removed by setting $\beta=0$, accuracy drops from 84.68\% to 73.88\%, and robustness drops from 50.57\% to 22.24\%. 
Without Diversity constraint, the policy network's training failed because the output policy distribution became concentrated on a few sub-policies, assigning zero probabilities to the remaining ones. 
The REINFORCE method could not recover from this situation because it no longer explored other options. 
This underscores the importance of maintaining a certain level of Diversity constraint in our policy learning. 
However, no clear benefit is observed as this constraint is further strengthened by raising $\beta$, as shown in {\cref{fig: prn beta}} and {\cref{fig: vit beta}}.

\subsubsection{Policy Model Architecture}
\label{app: ablation policy model}
Interestingly, we observed in \cref{tab: ablation policy model} that for CIFAR10 a relatively small model WideResNet10-1 (a WideResNets with depth 10 and widening factor 1) with 0.08M parameters is sufficient for learning the DA policy for a relatively large target model PRN18 with 11.17M parameters and further increasing capacity beyond this scale, even 100x, does not benefit either accuracy or robustness.
Therefore, the policy model can be much smaller than the target model.

\begin{table}[]
    \centering
    \caption{\textbf{Comparison of the various policy model backbone architectures} on CIFAR10 with a target model of PRN18.}
    \label{tab: ablation policy model}
    \begin{tabular}{@{}lccc@{}}
    \toprule
    \multicolumn{1}{c}{Model} & Size (M) & Clean       & AA          \\ \midrule
    WRN10-1                          & 0.08           & 84.16       & 50.25       \\
    WRN22-1                          & 0.27           & 84.32       & {\textbf{50.57}} \\
    WRN34-1                          & 0.47           & {\textbf{84.73}} & {\ul 50.38}       \\
    WRN70-1                          & {\ul 1.05}           & 84.04       & 50.28       \\
    PRN18 & {\textbf{11.17}} &  {\ul 84.68} & {\textbf{50.57}} \\ \bottomrule
    \end{tabular}
\end{table}

\begin{figure*}[!htb]
    \centering
    
    \subfloat[]{\includegraphics[width=.33\linewidth]{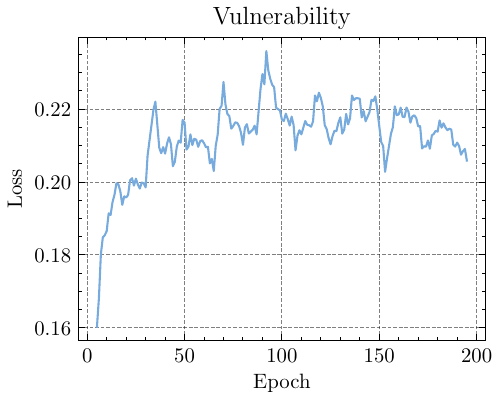}}
    \subfloat[]{\includegraphics[width=.33\linewidth]{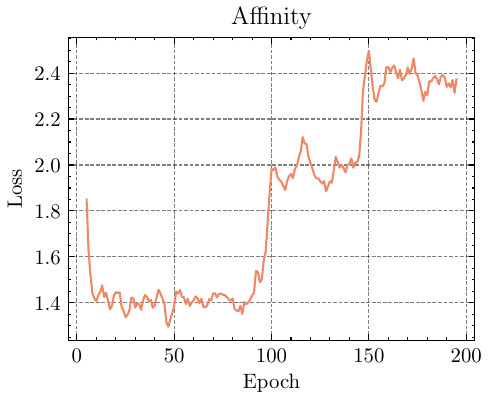}}
    \subfloat[]{\includegraphics[width=.33\linewidth]{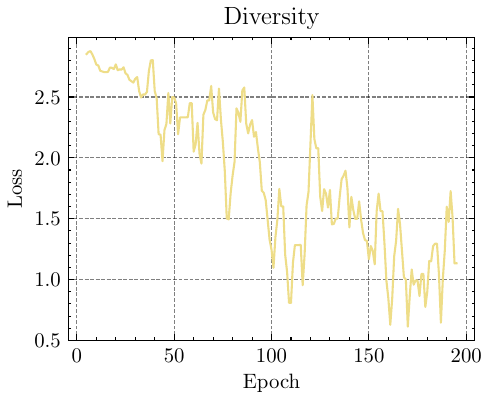}}
     
    \caption{\textbf{The progression of the three proposed policy learning objectives} throughout the AROID training process on CIFAR10 for WRN34-10.
    Lines are smoothed with a moving average over 5 epochs for improved clarity.}
    \label{fig: policy objective trajectory}
\end{figure*}

\subsubsection{Uniform Sampling}
\label{app: ablation uniform sampling}
We performed AT using data augmentations uniformly sampled from AROID’s data augmentation space. The results are labeled Uniform in \cref{tab: uniform sampling}. As shown in the table, AROID significantly improves accuracy and robustness over its uniformly sampled counterpart suggesting the necessity of optimizing the data augmentation policy.


\begin{table}[tb]
    \centering
    \caption{\textbf{Comparison of uniform sampling from AROID DA space} on CIFAR10 with PRN18.}
    \label{tab: uniform sampling}
    \begin{tabular}{lll}
    \toprule
    \multicolumn{1}{c}{DA} & \multicolumn{1}{c}{Clean}                   & \multicolumn{1}{c}{AA}                      \\ \midrule
    RandomCrop & {\ul 82.50} & 48.21       \\
    Uniform  & 81.00       & {\ul 49.18} \\
    AROID                  & \textbf{84.68} & {\textbf{50.57}} \\ 
    \bottomrule
    \end{tabular}%
\end{table}

\subsection{Analysis of Learned DA Policies} \label{app: da policies vis}
This section first analyzes the dynamics of the proposed policy learning objectives during training ({\cref{sec: dynamics of policy objectives}}).
It then visualizes the learned data augmentation policies sampled over a course of training ({\cref{sec: visualization of learning da policies}}). 
Last, it visualizes some image samples transformed by the learned data augmentation policies ({\cref{sec: visualization of augmented samples}}).

\subsubsection{Progression of Policy Learning Objectives}
\label{sec: dynamics of policy objectives}

To understand the dynamics of the learned data augmentation policy, \cref{fig: policy objective trajectory} visualizes the progression of the three proposed policy learning objectives throughout the AROID training process.
Generally, Vulnerability represents the adversarial vulnerability of the augmented data, Affinity reflects the distribution shift caused by data augmentation, and Diversity is negatively correlated with the diversity of data augmentation (lower Diversity implies greater diversity).
It is observed that during training, Vulnerability and Affinity increase while Diversity decreases.
These trends suggest that the data augmentation sampled from the learned policies becomes progressively harder, in terms of both adversarial vulnerability and distribution shift, and more diverse throughout the training process. 
This aligns with the goal of our policy learning as described in \cref{equ: overall policy learning objective} to encourage an increase in Vulnerability while regularizing Affinity and Diversity to decrease. 
It is important to note that an increase, rather than a decrease, is observed in the Affinity loss because Affinity was regularized with a decaying strength (in this case 0.4, 0.2, 0.1).

\subsubsection{Visualization of Learned DA Policies}
\label{sec: visualization of learning da policies}

\begin{figure}[!tb]
    \centering
    
    \subfloat[]{
        \includegraphics[width=.42\linewidth]{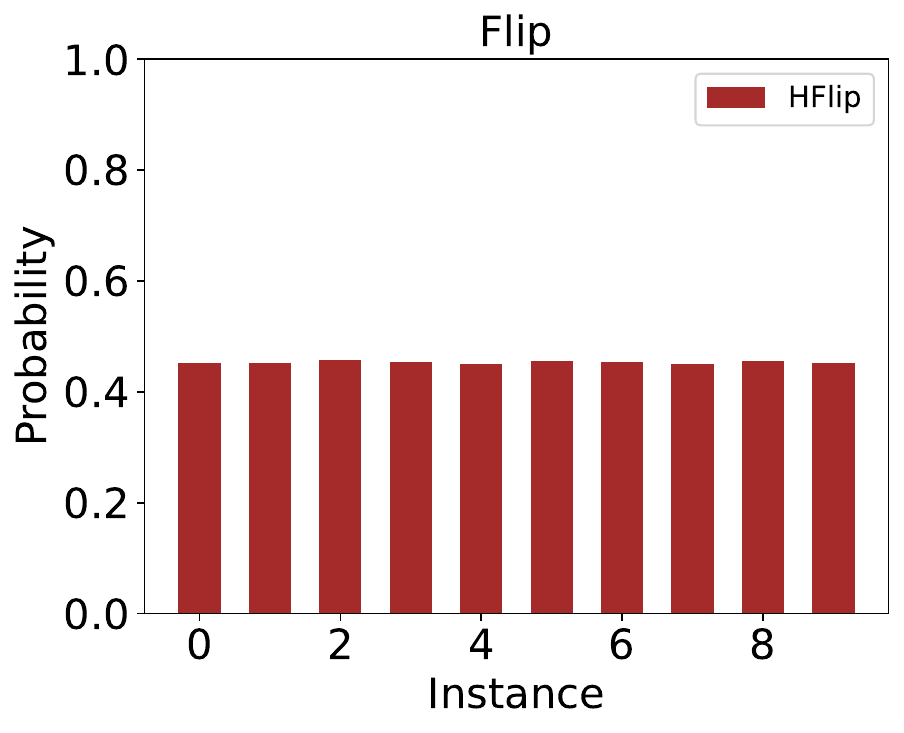}
        \label{fig: instance vis flip}
    }
    \subfloat[]{
        \includegraphics[width=.5\linewidth]{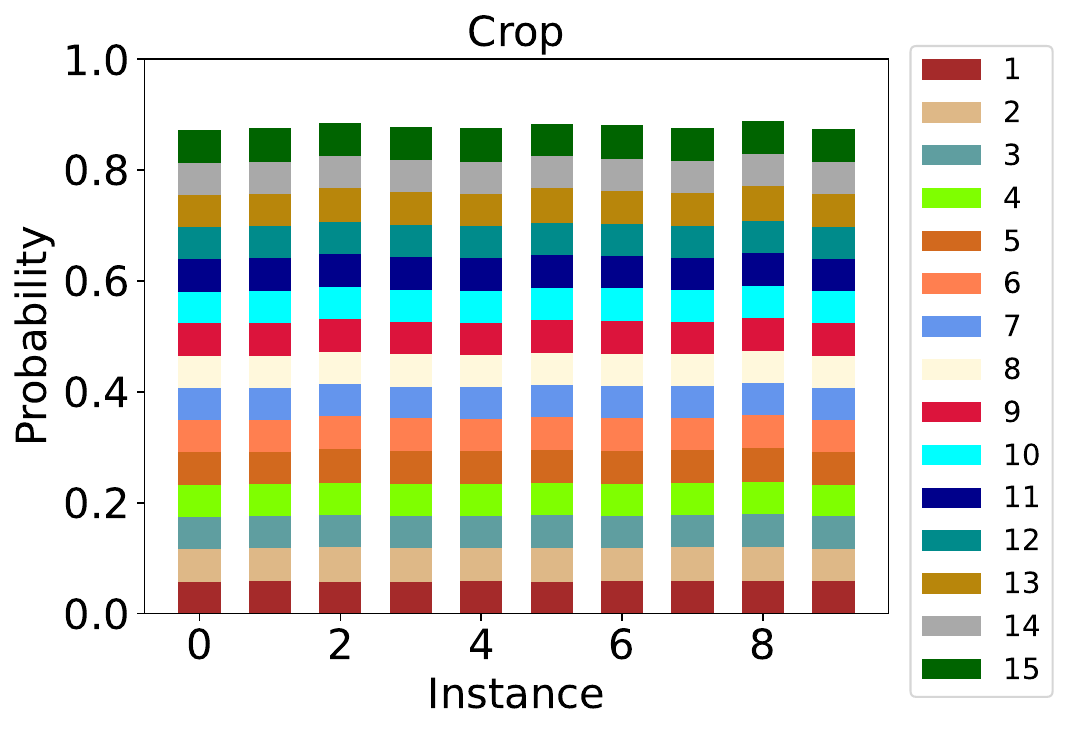}
        \label{fig: instance vis crop}
    }

    \subfloat[]{
        \includegraphics[width=.5\linewidth]{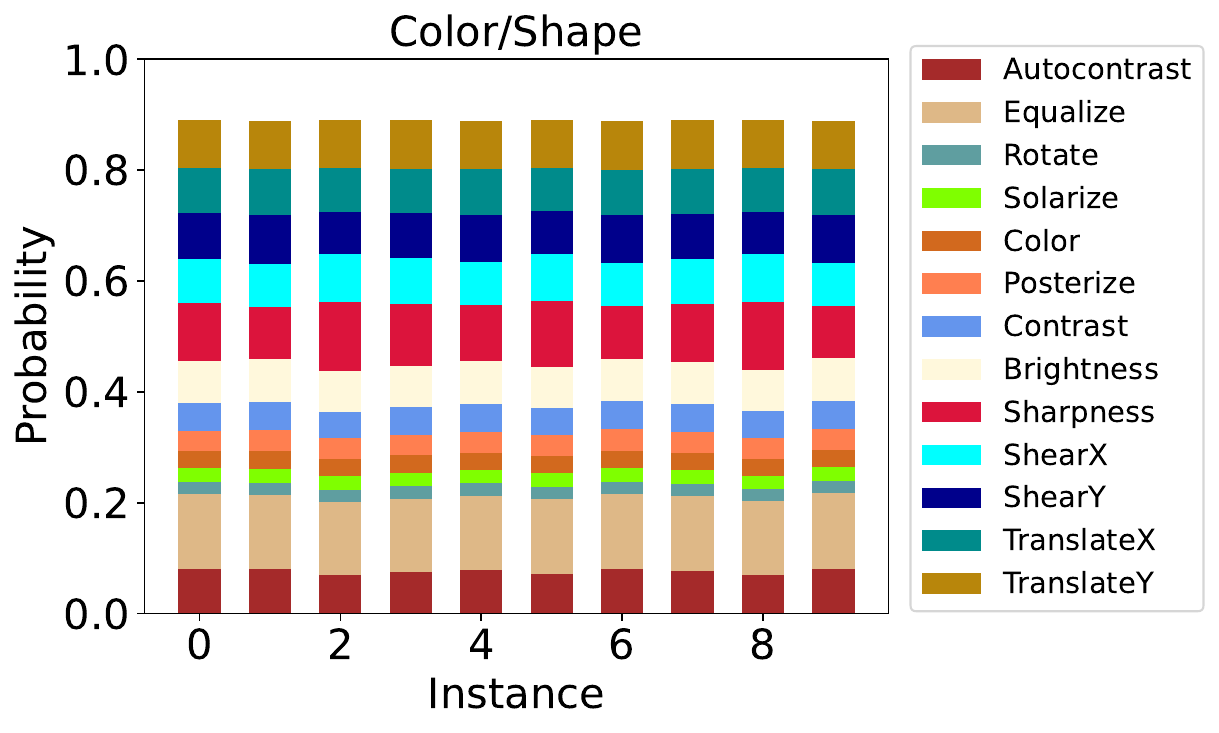}
        \label{fig: instance vis colsha}
    }
    \subfloat[]{
        \includegraphics[width=.45\linewidth]{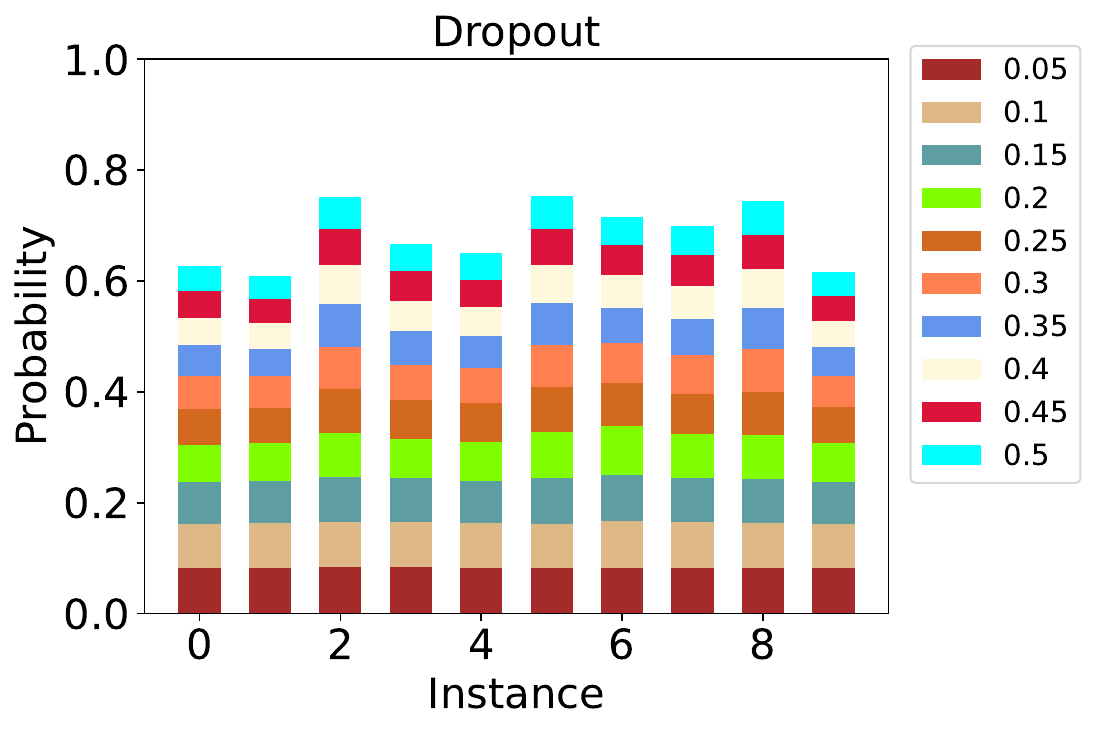}
        \label{fig: instance vis dropout}
    }

    \subfloat{
        \includegraphics[width=.15\linewidth]{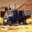}
    }
    \subfloat{
        \includegraphics[width=.15\linewidth]{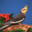}
    }
    \subfloat{
        \includegraphics[width=.15\linewidth]{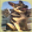}
    }
    \subfloat{
        \includegraphics[width=.15\linewidth]{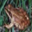}
    }
    \subfloat{
        \includegraphics[width=.15\linewidth]{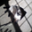}
    }

    \subfloat{
        \includegraphics[width=.15\linewidth]{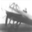}
    }
    \subfloat{
        \includegraphics[width=.15\linewidth]{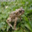}
    }
    \subfloat{
        \includegraphics[width=.15\linewidth]{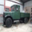}
    }
    \subfloat{
        \includegraphics[width=.15\linewidth]{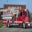}
    }
    \subfloat{
        \includegraphics[width=.15\linewidth]{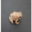}
    }
    
    \caption{\textbf{Visualization of the learned DA policies}, applied to ten images randomly sampled from CIFAR10 training set, for the Flip, Crop, Color/Shape and Dropout types of augmentations. The policy model is resumed from a checkpoint saved at the end of $110^{th}$ epoch when training a WRN34-10 model on CIFAR10 (following the training setting as specified in \cref{app: experiment setting}).
    The sampled ten images are visualized at the bottom in the order of the x-axis in the above bar-charts. The chance of applying no transformation (Identity) is the gap between the colored bar and the top (i.e., score of 1.0). In the Color/Shape group, the probabilities of different magnitudes are not shown separately, but are summed to get the overall probability of a transformation.}
    \label{fig: aug vis instance}
\end{figure}

\begin{figure}[!tb]
    \centering
    
    \subfloat[]{
        \includegraphics[width=.42\linewidth]{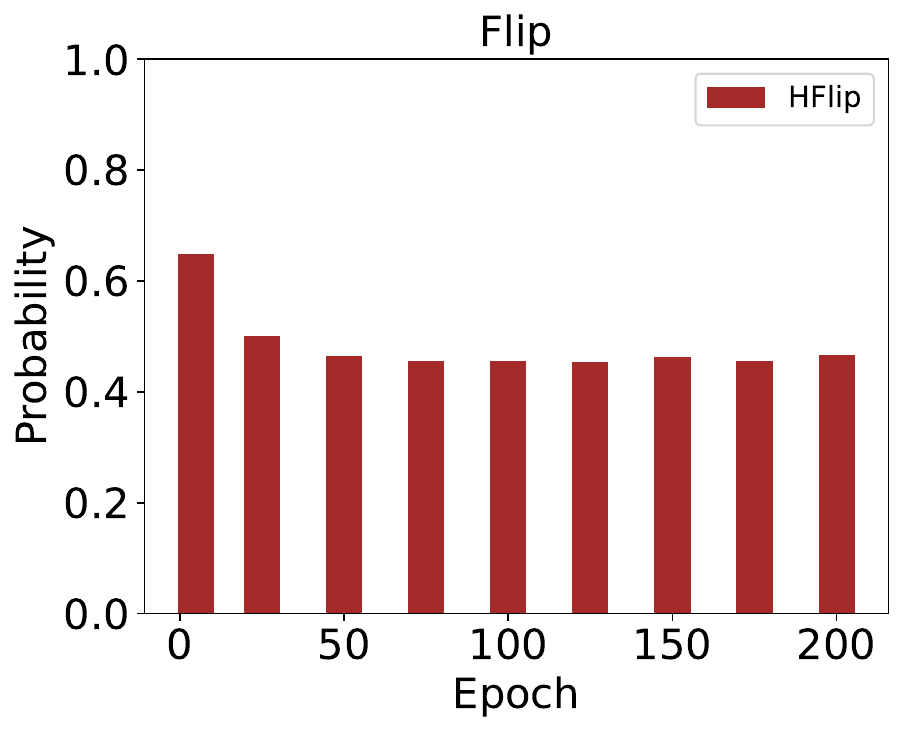}
        \label{fig: trajectory vis flip}
    }
    \subfloat[]{
        \includegraphics[width=.5\linewidth]{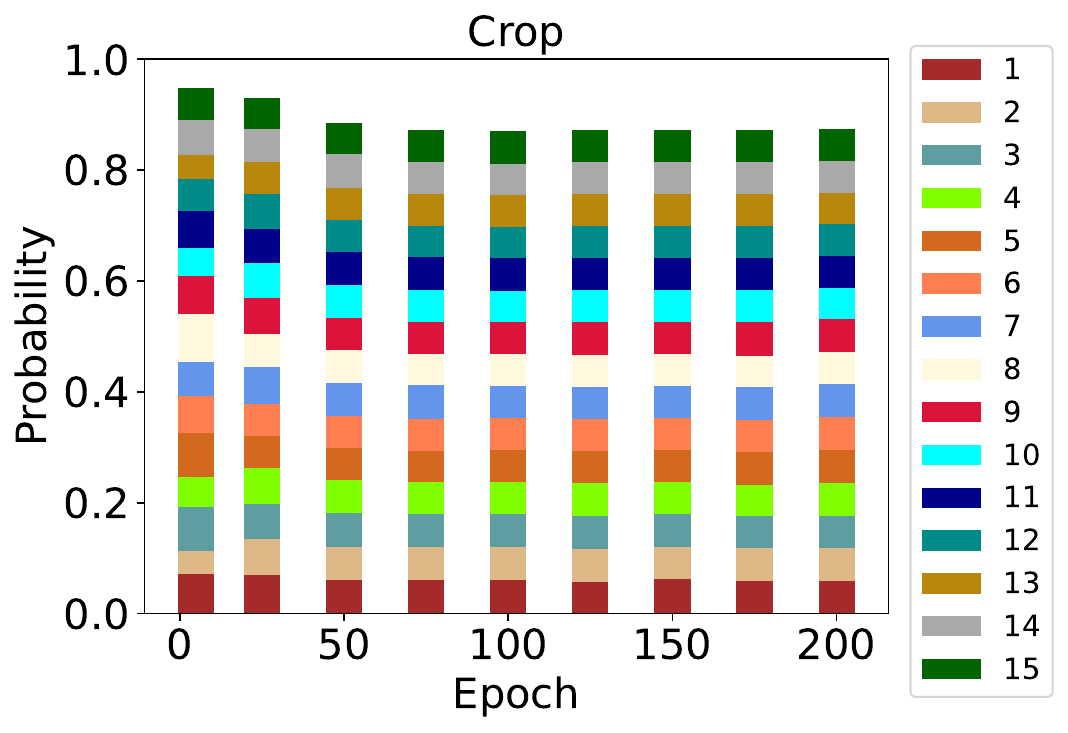}
        \label{fig: trajectory vis crop}
    }

    \subfloat[]{
        \includegraphics[width=.5\linewidth]{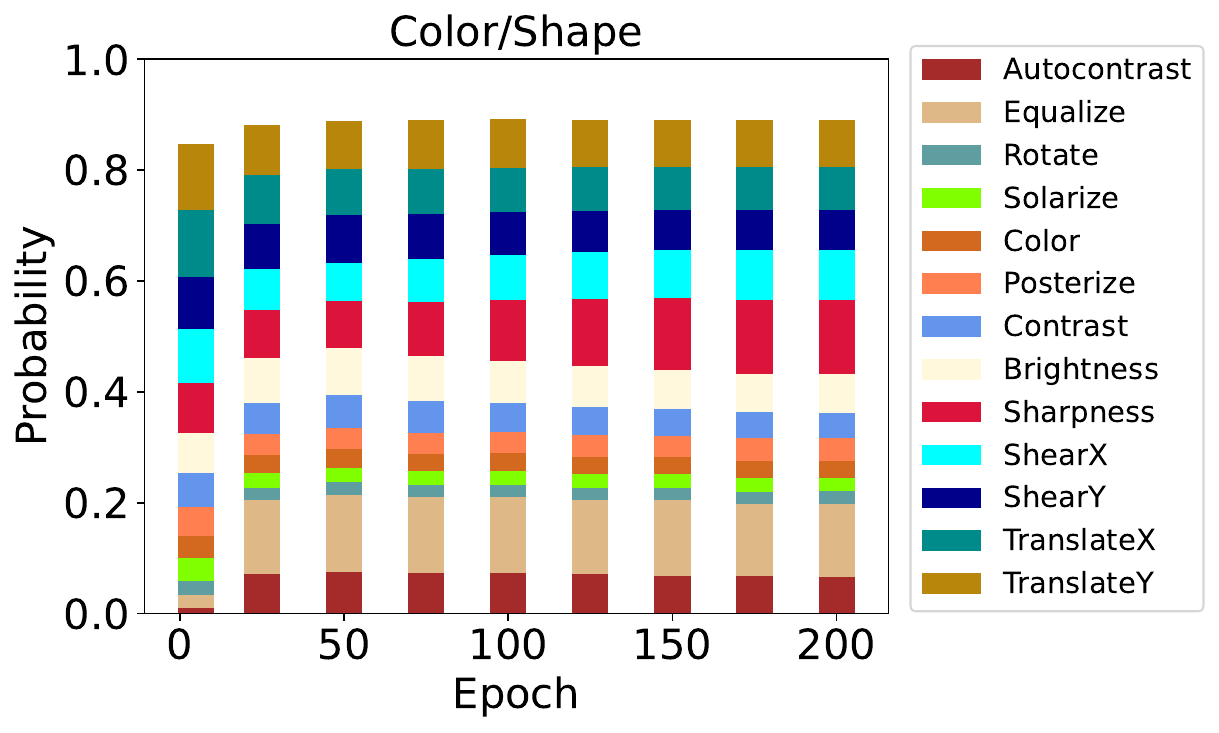}
        \label{fig: trajectory vis colsha}
    }
    \subfloat[]{
        \includegraphics[width=.45\linewidth]{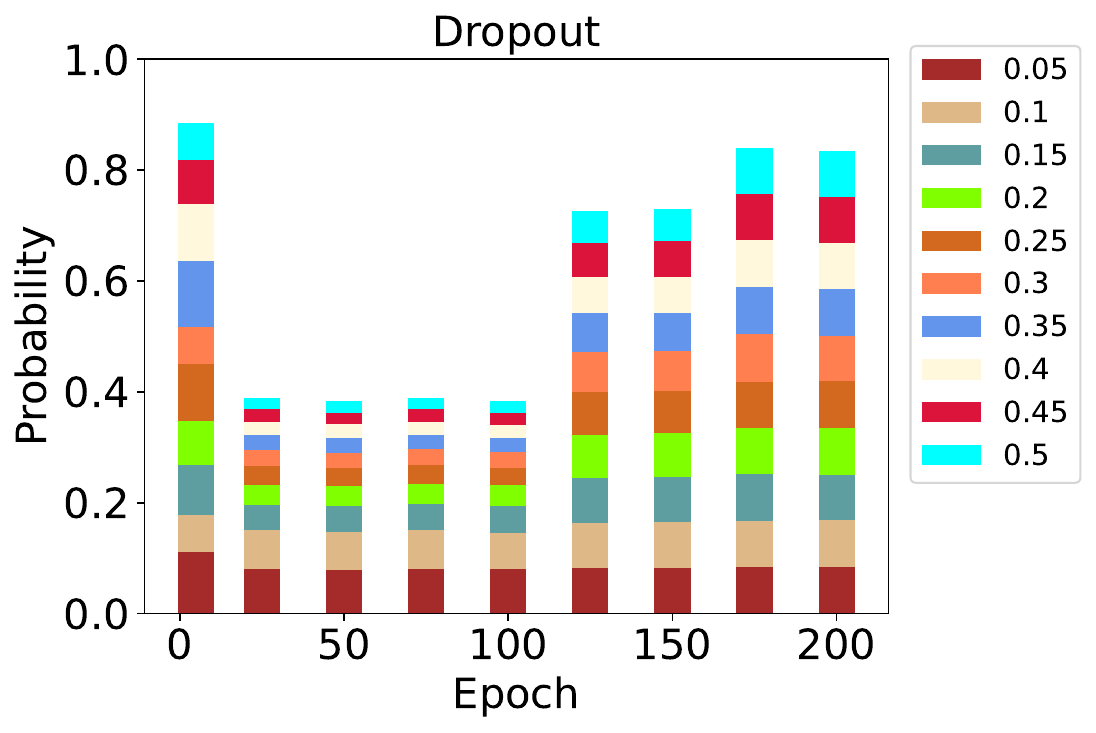}
        \label{fig: trajectory vis dropout}
    }

    \subfloat{
        \includegraphics[width=.15\linewidth]{0.png}
    }
    \hfill
    
    \caption{\textbf{Visualization of how the learned DA policies evolve as training progresses}. The same, randomly sampled, image (visualized at the bottom) was used across epochs (5, 25, 50, 75, 100, 125, 150, 175, 200) to produce the policies. The first bar in each sub-figure corresponds to the epoch 5 and describes the initial state of the policy model (training of policy model starts from epoch 5). For each bar in the figures, the policy model was resumed from the checkpoint saved at the corresponding epoch (x-axis) in the same course of training. The chance of applying no transformation (Identity) is the gap between the colored bar and the top (i.e., the score of 1.0). In the Color/Shape group, the probabilities of different magnitudes are not shown separately, but are summed to get the overall probability of a transformation.}
    \label{fig: aug vis trajectory}
\end{figure}

\begin{figure*}[!htp]
    \centering
    
    \subfloat{\includegraphics[width=.098\linewidth]{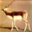}}
    \hfill
    \subfloat{\includegraphics[width=.098\linewidth]{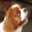}}
    \hfill
    \subfloat{\includegraphics[width=.098\linewidth]{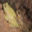}}
    \hfill
    \subfloat{\includegraphics[width=.098\linewidth]{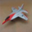}}
    \hfill
    \subfloat{\includegraphics[width=.098\linewidth]{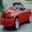}}
    \hfill
    \subfloat{\includegraphics[width=.098\linewidth]{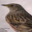}}
    \hfill
    \subfloat{\includegraphics[width=.098\linewidth]{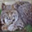}}
    \hfill
    \subfloat{\includegraphics[width=.098\linewidth]{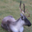}}
    \hfill
    \subfloat{\includegraphics[width=.098\linewidth]{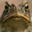}}
    \hfill
    \subfloat{\includegraphics[width=.098\linewidth]{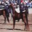}}

    \subfloat{\includegraphics[width=.098\linewidth]{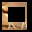}}
    \hfill
    \subfloat{\includegraphics[width=.098\linewidth]{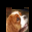}}
    \hfill
    \subfloat{\includegraphics[width=.098\linewidth]{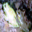}}
    \hfill
    \subfloat{\includegraphics[width=.098\linewidth]{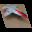}}
    \hfill
    \subfloat{\includegraphics[width=.098\linewidth]{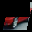}}
    \hfill
    \subfloat{\includegraphics[width=.098\linewidth]{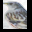}}
    \hfill
    \subfloat{\includegraphics[width=.098\linewidth]{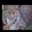}}
    \hfill
    \subfloat{\includegraphics[width=.098\linewidth]{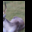}}
    \hfill
    \subfloat{\includegraphics[width=.098\linewidth]{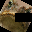}}
    \hfill
    \subfloat{\includegraphics[width=.098\linewidth]{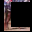}}

    \subfloat{\includegraphics[width=.098\linewidth]{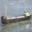}}
    \hfill
    \subfloat{\includegraphics[width=.098\linewidth]{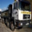}}
    \hfill
    \subfloat{\includegraphics[width=.098\linewidth]{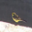}}
    \hfill
    \subfloat{\includegraphics[width=.098\linewidth]{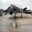}}
    \hfill
    \subfloat{\includegraphics[width=.098\linewidth]{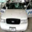}}
    \hfill
    \subfloat{\includegraphics[width=.098\linewidth]{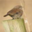}}
    \hfill
    \subfloat{\includegraphics[width=.098\linewidth]{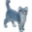}}
    \hfill
    \subfloat{\includegraphics[width=.098\linewidth]{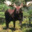}}
    \hfill
    \subfloat{\includegraphics[width=.098\linewidth]{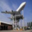}}
    \hfill
    \subfloat{\includegraphics[width=.098\linewidth]{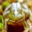}}

    \subfloat{\includegraphics[width=.098\linewidth]{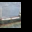}}
    \hfill
    \subfloat{\includegraphics[width=.098\linewidth]{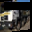}}
    \hfill
    \subfloat{\includegraphics[width=.098\linewidth]{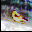}}
    \hfill
    \subfloat{\includegraphics[width=.098\linewidth]{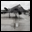}}
    \hfill
    \subfloat{\includegraphics[width=.098\linewidth]{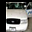}}
    \hfill
    \subfloat{\includegraphics[width=.098\linewidth]{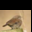}}
    \hfill
    \subfloat{\includegraphics[width=.098\linewidth]{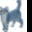}}
    \hfill
    \subfloat{\includegraphics[width=.098\linewidth]{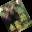}}
    \hfill
    \subfloat{\includegraphics[width=.098\linewidth]{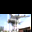}}
    \hfill
    \subfloat{\includegraphics[width=.098\linewidth]{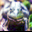}}

    \caption{\textbf{Visualization of 20 randomly-sampled pairs of original (odd rows) and augmented (even rows) samples} from CIFAR10. 
    The policy model is the same as that used for \cref{fig: aug vis instance}.}
    \label{fig: aroid augmented sampled visualization}
\end{figure*}

\cref{fig: aug vis instance} visualizes the learned distribution of DAs for different, randomly sampled, data instances. Instance-wise variation of the learned DA policy is visible for the Color/Shape augmentations (\cref{fig: instance vis colsha}) and evident for the Dropout augmentations (\cref{fig: instance vis dropout}), but subtle in the rest (\cref{fig: instance vis flip} and \cref{fig: instance vis crop}). 
Note that even for the different data instances from the same class (e.g., instances 4, 7, 10 from the class "frog"), the learned DA distributions can still differ considerably (\cref{fig: instance vis dropout}).
This confirms that (1) AROID is able to capture and meet the varied demand of augmentations from different data instances, and (2) such demand exists for some, but not all, augmentations. 
These observations may explain why many instance-agnostic DA methods such as IDBH, despite being inferior to ours, still work reasonably well (see \cref{tab: data augmentation robustness}). 

It was also observed in \cref{fig: aug vis trajectory} that the learned DA policy for the same data instance evolved as training progressed. In the Color/Shape group (\cref{fig: trajectory vis colsha}), augmentations like Sharpness became observably more likely to be selected while others such as ShearY became less probable as training continued. 
Dropout (i.e. Erasing; \cref{fig: trajectory vis dropout}) particularly with large magnitudes was rarely applied prior to 100th epoch, i.e., the first decay of learning rate. 
The possibility of applying Crop (i.e. Cropshift; \cref{fig: trajectory vis crop}) and Flip (i.e. HorizontalFlip; \cref{fig: trajectory vis flip}) first dropped until the first decay of learning rate and then stayed nearly constant afterwards. 

Consistent to the previous findings on ST \citep{cubuk_autoaugment_2019} and harmful augmentations \citep{rebuffi_data_2021}, we observed that AT on CIFAR10 favored mostly color-based augmentations like Equalize and Sharpness and disfavored geometric augmentations like Rotate and harmful augmentations like Solarize and Posterize (see both \cref{fig: instance vis colsha} and \cref{fig: trajectory vis colsha}). This verifies the effectiveness of our DA policy learning algorithm.

\subsubsection{Visualization of Augmented Data Samples}
\label{sec: visualization of augmented samples}

\cref{fig: aroid augmented sampled visualization} depicts 20 pairs of original and augmented data samples from CIFAR10. 
The visualization demonstrates that \textbf{our method effectively enhances the diversity of augmented data samples}. 
While the original and augmented data samples are paired here in a one-to-one manner, the learned policy enables the generation of a much larger variety of distinct augmented data.

\section{Conclusions}
This work introduces an approach, dubbed AROID, to efficiently learn online, instance-wise, DA policies for improved robust generalization in AT. 
AROID is the first automated DA method specifically for AT. 
Extensive experiments show its superiority over both alternative DA methods and contemporary AT methods in terms of accuracy and robustness.
AROID has also significantly reduces the cost of policy search making automated data augmentation practical to use for adversarial training, even for large datasets.
AROID can be also used in an offline mode to further save on computation.
The learned DA policies are visualized to verify the effectiveness of AROID and understand the preference of AT for DA. 

However, AROID has some limitations as well. First, despite being more efficient than IDBH, it still adds extra computational burden to training, unless AROID-T is used. This could harm its scalability to larger datasets and model architectures. Second, the Diversity objective enforces a minimal chance (set by the  lower limit) of applying harmful transformations and/or harmful magnitudes if they are included in the search space. This constrains the ability of AROID to explore a wider (less filtered) search space. Future works could investigate more efficient AutoML algorithms for learning DA policies for AT, and design new policy learning objectives to reduce the number of hyperparameters and alleviate the side-effect of Diversity.

\paragraph{Acknowledgments}
The authors gratefully acknowledge use of the King's Computational Research, Engineering and Technology Environment (CREATE) for carrying out the experiments described in this paper.
This work was funded by a scholarship from the King’s - China Scholarship Council (K-CSC).


\begin{appendices}

\section{DA Search Space} 
\label{app: augmentation space}

\cref{tab: augmentation space} shows the complete DA search space used by AROID. For Color/Shape group, we adopted the same operations as RandAugment's, but discretize the range of magnitudes for each operation into 10 even values if possible. For Erasing in Dropout group, the magnitude corresponds to the scale (the proportion of erased area against input image), while the aspect ratio (of erased area) is uniformly sampled from range $(0.3, 3.3)$. The search space only defines the operations and their magnitudes, while the probabilities of applying these operations are learned by AROID.

\begin{sidewaystable*}[]
\centering
\caption{Augmentation space}
\label{tab: augmentation space}
\begin{tabular}{@{}llllllllllll@{}}
\toprule
\multicolumn{3}{c}{Flip} & \multicolumn{3}{|c|}{Crop}            & \multicolumn{3}{|c|}{Color/Shape}                 & \multicolumn{3}{|c}{Dropout}             \\ \midrule
operations &
magnitudes &
  \multicolumn{1}{c|}{count} &
operations &
magnitudes &
  \multicolumn{1}{c|}{count} &
operations &
magnitudes &
  \multicolumn{1}{c|}{count} &
operations &
magnitudes &
  \multicolumn{1}{c}{count} \\ \midrule
Identity         & - & 1 & Identity  & -                  & 1  & Identity     & -                           & 1  & Identity & -           & 1  \\
Horiz. Flip      & - & 1 & Cropshift & 1, 2, 3, 4, 5,      & 15 & Autocontrast & -                           & 1  & Erasing  & .05, .10,   & 10 \\
                 &   &   &           & 6, 7, 8, 9,        &    & Equalize     & -                           & 1  &          & .15, .20,   &    \\
                 &   &   &           & 10, 11, 12,        &    & Posterize    & 4, 5, 6, 7, 8               & 5  &          & .25, .30,   &    \\
                 &   &   &           & 13, 14, 15         &    & Solarize     & 25, 51, 76, 102,         & 10 &          & .35, .40,   &    \\
                 &   &   &           &                    &    &              & 128, 153, 179,      &    &          & .45, .50    &    \\
                 &   &   &           &                    &    &              & 204, 230, 256             &    &          &              &    \\ 
                 &   &   &           &                    &    & Rotate       & 3, 6, 9, 12, 15,             & 10 &          &              &    \\
                 &   &   &           &                    &    &              & 18, 21, 24, 27, 30          &    &          &              &    \\
                 &   &   &           &                    &    & ShearX       & .03, .06, .09,      & 10 &          &              &    \\
                 &   &   &           &                    &    &              & .12, .15, .18,     &    &          &              &    \\
                 &   &   &           &                    &    &              &  .21, .24, .27, .30             &    &          &              &    \\ 
                 &   &   &           &                    &    & ShearY       & .03, .06, .09,      & 10 &          &              &    \\
                 &   &   &           &                    &    &              & .12, .15, .18,     &    &          &              &    \\
                 &   &   &           &                    &    &              &  .21, .24, .27, .30             &    &          &              &    \\ 
                 &   &   &           &                    &    & TranslateX   & 1, 2, 3, 4, 5,               & 10 &          &              &    \\
                 &   &   &           &                    &    &              & 6, 7, 8, 9, 10              &    &          &              &    \\
                 &   &   &           &                    &    & TranslateY   & 1, 2, 3, 4, 5,               & 10 &          &              &    \\
                 &   &   &           &                    &    &              & 6, 7, 8, 9, 10              &    &          &              &    \\
                 &   &   &           &                    &    & Color        & .28, .46, .64, .82,     & 10 &          &              &    \\
                 &   &   &           &                    &    &              & 1.0, 1.18, 1.36,  &    &          &              &    \\
                 &   &   &           &                    &    &              & 1.54, 1.72, 1.9             &    &          &              &    \\ 
                 &   &   &           &                    &    & Contrast     & .28, .46, .64, .82,     & 10 &          &              &    \\
                 &   &   &           &                    &    &              & 1.0, 1.18, 1.36,  &    &          &              &    \\
                 &   &   &           &                    &    &              & 1.54, 1.72, 1.9             &    &          &              &    \\ 
                 &   &   &           &                    &    & Brightness   & .28, .46, .64, .82,      & 10 &          &              &    \\
                 &   &   &           &                    &    &              & 1.0, 1.18, 1.36,  &    &          &              &    \\
                 &   &   &           &                    &    &              & 1.54, 1.72, 1.9             &    &          &              &    \\ 
                 &   &   &           &                    &    & Sharpness    & .28, .46, .64, .82,         & 10 &          &              &    \\
                 &   &   &           &                    &    &              & 1.0, 1.18, 1.36,            &    &          &              &    \\
                 &   &   &           &                    &    &              & 1.54, 1.72, 1.9             &    &          &              &    \\ \bottomrule
\end{tabular}%
\end{sidewaystable*}

\section{Derivation}
\label{app: detailed derivation}

This section discusses how we derive the gradients of Hardness metric w.r.t. the parameters of the policy model:
\begin{equation}
    \frac{\partial \mathbb{E}_{i \in B} \mathcal{L}_{hrd}(\bm{x}_i)}{\partial \bm{\theta}_{plc}}
    \label{equ: hardness gradient}
\end{equation}
First, we rewrite \cref{equ: hardness gradient} as below, so that we can focus on the gradient derivation part. 
\begin{equation}
    \frac{1}{B}\sum_{i=1}^B \frac{\partial \mathcal{L}_{hrd}(\bm{x}_i)}{\partial \bm{\theta}_{plc}}
    \label{equ: hardness gradient simplified}
\end{equation}
Next, to apply the REINFORCE algorithm, we substitute the gradient of the $\mathcal{L}_{hrd}$ for a sampled trajectory in \cref{equ: hardness gradient simplified} with the gradient of the expected $\mathcal{L}_{hrd}$ for multiple sampled trajectories as
\begin{equation}
    \frac{1}{B}\sum_{i=1}^B \frac{\partial \mathbb{E}_{t \in T} \mathcal{L}_{hrd}^{(t)}(\bm{x}_i)}{\partial \bm{\theta}_{plc}}
    \label{equ: hardness gradient expected}
\end{equation}
By applying the REINFORCE algorithm, we have (batch averaging is omitted for simplicity)
\begin{align}
    &\frac{\partial \mathbb{E}_{t \in T} \mathcal{L}_{hrd}^{(t)}(\bm{x}_i)}{\partial \bm{\theta}_{plc}} \nonumber \\
    &= \frac{\partial \sum_{t=1}^T \mathcal{P}_{(t)}(\bm{x}_i) \mathcal{L}_{hrd}^{(t)}(\bm{x}_i)}{\partial \bm{\theta}_{plc}} \\
    &= \sum_{t=1}^T \frac{\partial \mathcal{P}_{(t)}(\bm{x}_i)}{\partial \bm{\theta}_{plc}} \mathcal{L}_{hrd}^{(t)}(\bm{x}_i) \\
    &= \sum_{t=1}^T \mathcal{P}_{(t)}(\bm{x}_i) \frac{\partial \log(\mathcal{P}_{(t)}(\bm{x}_i))}{\partial \bm{\theta}_{plc}} \mathcal{L}_{hrd}^{(t)}(\bm{x}_i) \\
    &= \mathbb{E}_{i \in T} \frac{\partial \log(\mathcal{P}_{(t)}(\bm{x}_i))}{\partial \bm{\theta}_{plc}} \mathcal{L}_{hrd}^{(t)}(\bm{x}_i)
    \label{equ: hardness gradient REINFORCE}
\end{align}
$\mathcal{P}_{(t)}(\bm{x}_i)$ is the probability of sampled trajectory. Following the previous practices \citep{zhang_adversarial_2020, lin_online_2019, jia2022adversarial}, we approximate \cref{equ: hardness gradient REINFORCE} as
\begin{align}
    \frac{1}{T}\sum_{t=1}^T \frac{\partial \log(\mathcal{P}_{(t)}(\bm{x}_i))}{\partial \bm{\theta}_{plc}} \mathcal{L}_{hrd}^{(t)}(\bm{x}_i)
    \label{equ: hardness gradient REINFORCE approx}
\end{align}
Next, by expanding $\mathcal{P}_{(t)} = \prod_{h=1}^H p_{(t)}^h$, we have
\begin{align}
    &\frac{1}{T}\sum_{t=1}^T \frac{\partial \log(\prod_{h=1}^H p_{(t)}^h(\bm{x}_i; \bm{\theta}_{plc}))}{\partial \bm{\theta}_{plc}} \mathcal{L}_{hrd}^{(t)}(\bm{x}_i) \\
    &\approx \frac{1}{T}\sum_{t=1}^T \frac{\partial \sum_{h=1}^H\log(p_{(t)}^h(\bm{x}_i; \bm{\theta}_{plc}))}{\partial \bm{\theta}_{plc}} \mathcal{L}_{hrd}^{(t)}(\bm{x}_i)\\
    &\approx \frac{1}{T}\sum_{t=1}^T \sum_{h=1}^H \frac{\partial \log(p_{(t)}^h(\bm{x}_i; \bm{\theta}_{plc}))}{\partial \bm{\theta}_{plc}} \mathcal{L}_{hrd}^{(t)}(\bm{x}_i)
    \label{equ: hardness gradient REINFORCE expanded prob}
\end{align}
To reduce the variance of gradient estimation, we apply the baseline trick by subtracting mean value, $\tilde{\mathcal{L}_{hrd}}=\frac{1}{T}\sum_{t=1}^T\mathcal{L}_{hrd}^{(t)}(\bm{x}_i)$, from $\mathcal{L}_{hrd}^{(t)}$ as
\begin{align}
    \frac{1}{T}\sum_{t=1}^T \sum_{h=1}^H \frac{\partial \log(p_{(t)}^h(\bm{x}_i; \bm{\theta}_{plc}))}{\partial \bm{\theta}_{plc}} [\mathcal{L}_{hrd}^{(t)}(\bm{x}_i) - \tilde{\mathcal{L}_{hrd}}]
    \label{equ: hardness gradient REINFORCE baseline trick}
\end{align}
Eventually, by adding back the batch averaging, we have our ultimate form of gradients as
\begin{align}
    \frac{1}{B\cdot T}\sum_{i=1}^B\sum_{t=1}^T \sum_{h=1}^H \frac{\partial \log(p_{(t)}^h(\bm{x}_i))}{\partial \bm{\theta}_{plc}} [\mathcal{L}_{hrd}^{(t)}(\bm{x}_i) - \tilde{\mathcal{L}_{hrd}}]
    \label{equ: hardness gradient ultimate}
\end{align}

\section{Efficiency Analysis}
\label{app: efficiency policy learning}

The efficiency of AROID is analyzed here. $F_t$/$F_p$/$F_a$ and $B_t$/$B_p$/$B_a$ denote the cost of forward and backward pass on target/policy/affinity model respectively.
For each iteration of updating policy model, the major overhead is
\begin{itemize}
    \item Predict DA distribution: 1 $F_p$
    \item Vulnerability: for each of $T$ trajectories, 2 $(F_t + B_t)$ to generate adversarial examples and 1 $F_t$ to calculate loss. Overall, $(3F_t+2B_t)T$
    \item Affinity: 1 $F_a$ to calculate the loss of original data which is shared by all $T$ trajectories. 1 $F_a$ to calculate the loss of augmented data for each of $T$ trajectories. Overall, $(F_aT+F_a)$
    \item Diversity: the calculation of diversity loss adds negligible overhead and does not require $F$ or $B$
    \item Update policy model: 1 $B_p$
\end{itemize}
To sum up, one iteration of policy update costs
\begin{equation}
    (3F_t+2B_t)T+ (F_aT+F_a)+F_p+B_p
\end{equation}
Policy model is updated every $K$ iterations of target model, so the averaged policy learning cost per iteration of target model training is 
\begin{equation}
    [(3F_t+2B_t)T+ (F_aT+F_a)+F_p+B_p]/K
\end{equation}
The overall overhead of AROID is learning cost plus 1 $F_p$ for every iteration of target model to sample DA, so 
\begin{equation}
    [(3F_t+2B_t)T+ (F_aT+F_a)+F_p+B_p]/K+F_p
\end{equation}
In worst case, policy and affinity models use the same architecture as target model, so the cost is 
\begin{equation}
    [(4T+2)/K+1]F_t + (2T+1)B_t / K
\end{equation}
The most expensive setting we use is $T=8$ and $K=5$, so it costs $7.8F_t+3.4B_t$ roughly, assuming $2F_t=1B_t$, $4.8(F_t+B_t)$ in addition to $11(F_t+B_t)$ of underlying PGD10 AT. Overall, in worst case, AROID adds about 43.6\% extra computation to baseline AT. For a cheaper setting $T=4$ and $K=20$, the overhead is roughly $1.9F_t+0.45B_t$ about 10\% more than baseline AT.

\section{Experimental Set-ups} 
\label{app: experiment setting}

\subsection{Configuration of AROID}
Vulnerability objective was calculated based on PGD2 with a step size of 2/255 except that PGD1 with a step size of 4/255 for ImageNet.
The affinity models used the same architecture as the target model. 
The affinity models were pre-trained using ST with the same settings as their AT trained counterparts yet with no augmentation. 
Early stopping was used if training accuracy was close to 100\%. 
The policy model was trained using SGD with a constant learning rate (0.001 by default while 0.1 for Imagenette due to the reduced number of training epochs) and the same momentum as the target optimizer's. Gradient clipping was applied to stabilize the training of the policy model. 
In the initial five epochs of training, we did not train the policy model nor apply it to augment the data (no augmentation at all was applied) since the target model changed rapidly. 

\subsection{Configuration of Compared DA Methods}
\label{app: config of da methods}
AutoAugment was parameterized as in \citet{cubuk_autoaugment_2019} since we did not have sufficient resource to optimize.
For AutoAugment, augmentations were applied in the order of HorizontalFlip-RandomCrop-AutoAugment-Cutout (16x16)as in \citet{cubuk_autoaugment_2019}.
TrivialAugment is parameter-free so no tuning was needed.
For TrivialAugment, augmentations were applied in the order of HorizontalFlip-RandomCrop-TrivialAugment-Cutout (16x16) as in \citep{muller2021trivialaugment}.
For CutMix, $\alpha=0.25$ and $\beta=1$ on CIFAR10 as optimized in \citep{li_data_2023}; $\alpha=1$ and $\beta=1$ on Imagenette as suggested in \citep{yun2019cutmix}. For Cutout, the size of cut-out area was 20x20 on all three datasets as in \citep{li_data_2023}. 
Cutout and CutMix were applied with the default (baseline) augmentations in the order of HorizontalFlip-RandomCrop-Cutout and -CutMix respectively on CIFAR10 and Imagenette.
For IDBH, IDBH[strong]-CIFAR10 was used.

We only compare our method against the baseline and AutoAugment on ImageNet. AutoAugment is selected because it is one of the two methods closest to AROID and has a pre-optimized version for ImageNet while the other closest work IDBH doesn't. Due to the tremendous cost of conducting AT on ImageNet and the limit of our computational resource, we can't optimize other DA methods for AT on ImageNet so they are not included to avoid unfair comparison. In fact, like most other researchers, we don't have enough time and resource to train all competitive DA methods even without re-optimization of hyperparameters.

\subsection{Configuration of Compared State-of-the-art Robust Training Methods}
We only re-implemented the algorithms of SWA and AWP to report the result based on our runs, while the result of the others including MART, MART-AWP, SEAT, LAT-AT and LAS-AWP were copied directly from their original works except that the result of MART was copied from \citep{wu_adversarial_2020} for a better aligned training setting. 
SWA was implemented as in \citep{rebuffi_data_2021} with a decay rate of $\tau = 0.999$. AWP was configured as in \citep{wu_adversarial_2020} with $\beta = 0.005$. Note that the same configurations of SWA and AWP were used to train with baseline DA and AROID.




\end{appendices}

\bibliography{references,reference}

\end{document}